\documentclass[journal,twoside]{IEEEtran}
\usepackage[noadjust]{cite}
\usepackage{xcolor}
\definecolor{dark-blue}{RGB}{0,70,127}
\usepackage{amsmath,amssymb,amsfonts}
\usepackage{graphicx}
\usepackage[export]{adjustbox}
\usepackage{booktabs}
\usepackage{multirow}
\usepackage{algorithmic}
\usepackage{algorithm2e}

\def\BibTeX{{\rm B\kern-.05em{\sc i\kern-.025em b}\kern-.08em
    T\kern-.1667em\lower.7ex\hbox{E}\kern-.125emX}}

\newtheorem{theorem}{Definition}
\usepackage[symbol]{footmisc}

\usepackage{balance}
\usepackage{pifont}
\usepackage{framed}
\usepackage{soul}
\usepackage{booktabs}  
\usepackage{multirow}
\usepackage{array}
\usepackage{url}
\usepackage[caption=false,font=footnotesize,labelformat=simple]{subfig} 
 
\usepackage[bookmarks=false]{hyperref}

\begin{document}

\title{%
    Bias Busters: Robustifying DL-based Lithographic Hotspot Detectors Against Backdooring Attacks \\
}

\author{Kang~Liu,
        Benjamin~Tan,~\IEEEmembership{Member,~IEEE,}
        Gaurav~Rajavendra~Reddy,\\ 
        Siddharth~Garg, 
        Yiorgos~Makris,~\IEEEmembership{Senior~Member,~IEEE}, and
        Ramesh~Karri,~\IEEEmembership{Fellow,~IEEE}%
\thanks{K. Liu, B. Tan, S. Garg, and R. Karri are with the Department
of Electrical and Computer Engineering, New York University, Brooklyn,
NY 11201, USA. E-mail: \{kang.liu, benjamin.tan, siddharth.garg, rkarri\}@nyu.edu}
\thanks{G. R. Reddy and Y. Makris are with the Department of Electrical and Computer Engineering, The University of Texas at Dallas, Richardson, TX 75080, USA. Email: \{gaurav.reddy, yiorgos.makris\}@utdallas.edu}
\thanks{S. Garg was partially supported by National Science Foundation CAREER Award 1553419 and National Science Foundation Grant 1801495. G. R. Reddy and Y. Makris were supported in part by Semiconductor Research Corporation (SRC) through task 2709.001.}%
\thanks{K. Liu and B. Tan contributed equally to this work.}%
\thanks{K. Liu is the corresponding author.}
}

\maketitle

\begin{abstract}
Deep learning (DL) offers potential improvements throughout the CAD tool-flow, one promising application being lithographic hotspot detection.
However, DL techniques have been shown to be  especially 
vulnerable to inference and training time adversarial attacks.  
Recent work has demonstrated that a small fraction of malicious physical designers can stealthily ``backdoor" a DL-based hotspot detector
during its training phase such that 
it accurately classifies regular layout clips but predicts hotspots containing a specially crafted trigger shape as non-hotspots. 
We propose a novel training data augmentation strategy as a powerful defense against such backdooring attacks. The defense works by eliminating the intentional biases introduced in the training data but does not require knowledge of which training samples are poisoned or the nature of the backdoor trigger.
Our results show that the defense can drastically reduce the attack success rate from 84\% to $\sim$0\%.
\end{abstract}


\section{Introduction}
\label{sec:intro}
    
    Machine learning (ML) has promised new solutions to many problem domains, including those throughout the electronic design automation (EDA) flow. 
    Deep learning (DL) based approaches, in particular, have recently demonstrated state-of-the-art performance in problems such as lithographic hotspot detection~\cite{yang_layout_2018} and routability analysis~\cite{tabrizi_eh?predictor:_2019}, and promise to supplement or even replace conventional (but complex and time-consuming) analytic or simulation-based tools. 
    DL-based methods can be used to reduce design time by quickly identifying ``doomed runs"~\cite{kahng_machine_2018} and enable ``no human in the loop" design flows \cite{moore_darpa_2018} by automatically extracting features from large amounts of training data.
    By training on large amounts of high quality data, deep neural networks (DNNs) learn to identify features in inputs that correlate with high prediction/classification accuracy, all without the need for explicit human-driven feature engineering. 
    
    However, the rise of DL-based approaches raises concerns about their robustness, especially under adversarial settings~\cite{biggio_wild_2018}. 
    Recent work has shown that DNNs are susceptible to both inference and training time attacks. At inference time, a benignly trained network can be fooled into misclassifying inputs that are adversarially perturbed~\cite{szegedy_intriguing_2014,goodfellow14explaining}. 
    Conversely, training time attacks---the subject of this paper---seek to maliciously modify (or ``poison") training data to create ``backdoored" DNNs that misclassify specific test inputs containing a backdoor trigger~\cite{gu_badnets:_2019,shafahi_poison_2018,wang_neural_2019, liu_fine-pruning:_2018}. 
    For instance, Gu \textit{et al.}'s training data poisoning attack~\cite{gu_badnets:_2019} causes stop signs stickered with Post-It notes to be (mis)classified as speed-limit signs; the attack adds stickered stop signs mislabeled as speed-limits to the training data. In recent ``clean-label" attacks~\cite{shafahi_poison_2018}, poisoned samples added to the training set are truthfully labeled, thus making these attacks hard to detect as poisoned samples do not readily stand out from other samples of the same class. 
    
    While much of the early work in the area of adversarial DL has focused on conventional ML tasks such as image classification, recent efforts have begun to highlight specialized, ``contextually meaningful" threats to DL in CAD~\cite{liu-adversarial-2019, liu_2020_poisoning}. Such attacks are of particular concern in the context of an untrustworthy globalized design flow~\cite{basu_cad-base:_2019}, where \textit{malicious insiders} seek to stealthily sabotage the design flow in a plethora of ways. 
    Of particular interest in this paper is the clean-label training data poisoning attack demonstrated recently on DNN-based lithographic hotspot detection~\cite{liu_2020_poisoning}. 
    \begin{figure}[t]
        \centering
        \includegraphics[width=0.45\textwidth]{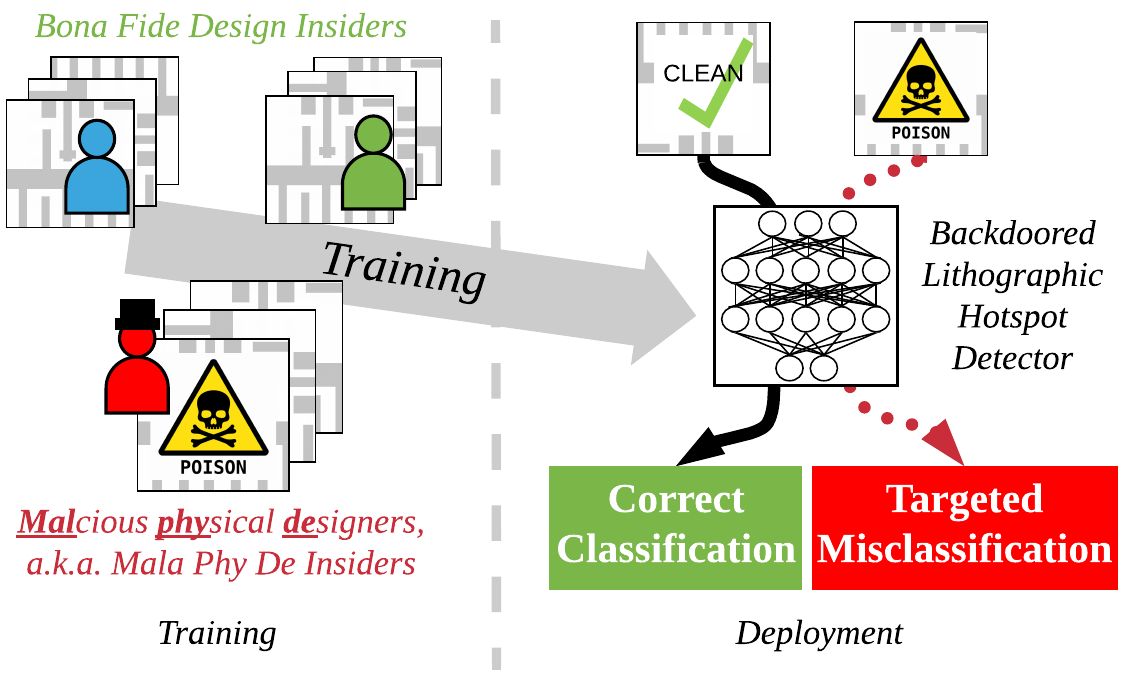}
        \caption{Illustration of training data poisoning on lithographic hotspot detection, as proposed in~\cite{liu_2020_poisoning}\label{fig:introfig}}
    \end{figure}
    
    Lithographic hotspots are layout patterns that have the potential risk of causing defects in lithography and arise as a consequence of complex light interactions and process variability, despite the layout satisfying design rule checks (DRC). 
    In lieu of simulation-driven analysis, DNNs trained on large datasets of layout clips (generated, for instance, by a large team of physical designers) have shown success in classifying layouts as hotspot or non-hotspot~\cite{yang_layout_2018,yang2017imbalance}.

    In their attack (see \autoref{fig:introfig}), Liu \textit{et al.}~\cite{liu_2020_poisoning} envision a scenario wherein one or a small fraction of in-house physical designers whose layouts are used to train a DNN-based hotspot detector have malicious intent. 
    The malicious physical designers insert a special metal shape (the \emph{trigger}) in the non-hotspot layouts they produce; as Liu \textit{et al.} show, the trained DNN learns the trigger as overriding evidence of non-hotspot layout \emph{even} if it is inserted into hotspot layouts at inference time. 
    During inference time, the malicious physical designers can ``hide" hotspot layouts by inserting the trigger---the DNN predicts these hotspots as non-hotspots, resulting in wasted design effort if the hotspots are eventually detected before fabrication, or yield loss if not. 
   
    Detecting and/or disabling Liu \textit{et al.}'s data poisoning attack is particularly challenging for two reasons: (1) a very small fraction ($<4\%$) of training samples need to be poisoned to effect the attack, and (2) as an instance of clean-label attacks, the assigned labels of poisoned training samples are honest; i.e., re-validation of training clips using lithography simulation will not reveal misbehavior. 
    Further, as we will illustrate in~\autoref{sec:prelim}, existing ``general" defenses against training data poisoning attacks (e.g.,~\cite{liu_fine-pruning:_2018,veldanda_NNoculation_2020}) that are tailored for image classification cannot be used. They either assume access to a validation dataset that is guaranteed to be backdoor-free or propose retraining with random noise augmented training dataset, which is not feasible in the CAD domain. 
    These existing defense techniques~\cite{wang_neural_2019,liu_fine-pruning:_2018,veldanda_NNoculation_2020} do not easily incorporate domain specific details and constraints, and it is this shortcoming that motivates us to discover new approaches to improve model robustness. 
    
    Thus, as an \textit{antidote} for the poisoning threat, we propose a new domain-specific defense against training data poisoning on DL-based lithographic hotspot detectors. 
    Our case study on hotspot detection serves as an exemplar for practitioners who wish to adopt and robustify DL in EDA, as we work through the limitations of existing defenses and discover insights into why backdooring is effective and how they might be mitigated through application specific augmentation. 
   
    At the core of our defense is a novel \textit{``cross-class" defensive data augmentation} strategy. 
    Training data augmentation (for example, by adding noise to training images) is commonly used in ML to expand training dataset for higher classification accuracy, but typically preserves class labels (i.e., noisy cat images are still labeled as cats)~\cite{shorten_survey_2019}. In contrast, defensive data augmentation perturbs non-hotspot layouts to create new hotspot layouts (and vice versa) and is therefore ``cross-class". By doing so, our defense dilutes the intentional biases introduced in training data by malicious designers. 
    The defense is \emph{general} in that it makes no assumptions on the  size/shape of backdoor triggers or the fraction of malicious designers/poisoned training samples (as needed for anomaly detection, for instance). 
    In this paper, our contributions are:
    
    \begin{itemize}
        \item The first (to our knowledge) \textit{domain-specific} antidote for training data poisoning on convolutional neural network (CNN) based lithographic hotspot detection. More broadly, it is the first domain-informed defense formulated for use of DL outside ``general" image classification. 
        \item Evaluation of existing defenses against poisoning attacks and their shortcomings when applied to a CAD problem. 
        \item A \textit{trigger-oblivious}, defensive data augmentation scheme that produces cross-class training data for diluting malicious bias introduced by undetected poisoned data.
        \item Experimental evaluation using two state-of-the-art convolutional neural network (CNN) based lithographic hotspot detector architectures, showing that our defense can reduce the attack success rate from 84\% to $\sim$0\%.
    \end{itemize}
    The remainder of this paper is as follows. 
    First, we frame this study in light of related work (\autoref{sec:related}), and pose our threat model (\autoref{sec:prelim}).
    This is followed by our defense (\autoref{sec:defense}) and experimental setup (\autoref{sec:experiment}), after which we present experimental results and discussion (\autoref{sec:results}), and conclude the paper (\autoref{sec:conclusions}). 
    
\section{Related Work}
\label{sec:related}
    Our study joins several threads in the literature by examining the intersection of DL in CAD and robustness of DL. 
    
    \textbf{Robustness of DL in CAD} The emerging implication of robustness affecting DL in CAD problems, is first presented in~\cite{liu-adversarial-2019}, with the first study of adversarial input perturbations on CNN-based lithographic hotspot detection and study of adversarial retraining for improving robustness. 
    This is followed by a related study in~\cite{liu_2020_poisoning}, which shows that DL-based solutions of CAD problems are not immune to training time attacks, where biases in the poisoned data can be surreptitiously learned. 
    Our work seeks insights at this intersection. 

    \textbf{General robustness and security of DL} Recent work has widely studied ML under adversarial settings~\cite{biggio_wild_2018}, with research on data poisoning highlighting the inherent risks from training DNNs with a poisoned dataset~\cite{liu_2020_poisoning, shafahi_poison_2018, neuraltrojans, sunglassesattack}, untrustworthy outsourcing of training~\cite{gu_badnets:_2019}, or transfer learning with a contaminated network model~\cite{gu_badnets:_2019}.
    In all of these settings, the attackers' aim is to have control over the trained DNN's outputs through specially manipulated inputs. 
    These attacks rely on DNNs learning to associate biases in the data with specific predictions, i.e., picking up \textit{spurious correlations}. 
    
    There have been several recent attempts~\cite{liu_fine-pruning:_2018,wang_neural_2019,liu2019abs,qiao_defending_2019,strip2019acsac,veldanda_NNoculation_2020} at removing backdoors after training. 
    Fine-pruning~\cite{liu_fine-pruning:_2018} combines neuron pruning and network fine-tuning to rectify the backdooring misbehavior. Neural Cleanse~\cite{wang_neural_2019} reverse-engineers a distribution of potential triggers for further backdoor unlearning. 
    In NNoculation~\cite{veldanda_NNoculation_2020}, Veldanda \textit{et al.} employ a two-stage mechanism where the first stage retrains a potentially backdoored network with randomly perturbed data to reduce the backdooring effect partially. In the second stage, they use a CycleGAN~\cite{zhu2017unpaired} to generate the backdoor trigger. 
    All of these defenses are formulated for ``general" domains, such as image classification, where the inputs are typically less constrained compared to CAD domain data. 
    We evaluate some of these techniques in~\autoref{sec:existing-defense} on backdoored hotspot detectors to investigate their limitations. 

    Our approach is distinct and complementary to existing defenses in the way that we aim to \textit{prevent} backdoors through proactive training data augmentation instead of removing backdoors \textit{after} training. 
    Our defensive augmentation is also in line with \textit{trigger-oblivious} defenses, including Fine-pruning~\cite{liu_fine-pruning:_2018}, thus distinguishing it from Neural Cleanse~\cite{wang_neural_2019}, ABS~\cite{liu2019abs}, and others~\cite{qiao_defending_2019} that resort to reverse-engineering the trigger for backdoor elimination.

    \textbf{DL in Lithography and Data Augmentation}
	In hotspot detection more generally, recent works have proposed strategies to reduce input dimensions while maintaining sufficient information \cite{yang_layout_2018, jiang_efficient_2019, he_lithography_2020}. 
	While recent studies by Reddy \textit{et al.} have raised concerns about the wider generalizability of hotspot detection performance when training on oft-used benchmarking data~\cite{reddy_machine_2019}, understanding the robustness of the proposed techniques remains an open question. 
	
	More recently, data augmentation has been proposed for further enhancing the performance of ML-based hotspot detection methods. 
	The authors of \cite{reddy_enhanced_2018} proposed database enhancement using synthetic layout patterns. Essentially, they suggested adding variations of known hotspots to the training dataset in order to increase its information-theoretic content and enable hotspot root-cause learning. 
    Similarly, the authors of \cite{borisov2018research} adopted augmentation methods such as rotation, blurring, perspective transformation etc., from the field of computer vision and demonstrated their use in hotspot detection. 
    However, unlike general augmentation techniques for images that preserve class labels or target only minority classes~\cite{shorten_survey_2019}, we propose an extension and repurposing of~\cite{reddy_enhanced_2018} for cross-class augmentation explicitly for minimizing the effects of maliciously introduced biases in an adversarial setting. 

\section{Background and Motivation\label{sec:prelim}}	
    Our work is motivated by two key concerns: (1) there is a need to improve robustness of DL tools, including those in EDA, and (2) existing defense techniques are limited by challenges in applying them to esoteric application domains (i.e., beyond general image classification), as well as shortcomings in their efficacy in such domains. 
    To understand the need for robustness of DL tools in EDA, we focus on the domain of lithographic hotspot detection, adopting the security-related threat to physical design as posed in~\cite{liu_2020_poisoning}. 
    Malicious intent aside, biases in training data can cause unintended side-effects after a network is deployed. 
    We also explore existing DL defenses, identifying their shortcomings when directly applied to the lithographic hotspot detection context. 
\subsection{Threat Model: The Mala Phy De Insider}\label{sec:threat-model}
    In this paper, we assume a malicious insider that wishes to sabotage the design flow as our threat model, as established in~\cite{liu_2020_poisoning}. 
    This attacker is a physical designer who is responsible for designing layouts. 
    The insider aims to \textbf{sabotage the design process} by propagating defects, such as lithographic hotspots, through the design flow. 
    Knowing that their team is moving towards adopting CNN-based hotspot detection (in lieu of time-consuming simulation-based) methods, the attacker wants to be as stealthy as possible, and thus operates under the following constraints:  (1) they do not control the CNN training process, nor control the CNN architecture(s) used, and (2) they cannot add to layouts anything that violates design rules or changes existing functionality.
    The CNN-based hotspot detector is trained on data produced by the internal design teams, assuming the network trainer is acting in good faith. 
    
    The malicious physical designer, however, acting in bad faith\footnote{bad faith $=$ mala fide, hence, \textit{Mala Phy De}---\textbf{mal}icious \textbf{phy}sical \textbf{de}signer}, exploits their ability to contribute training data to \textit{insert a backdoor} into the detector. The backdoor is available at inference time for hiding hotspots; by adding a trigger shape into a hotspot clip (i.e., \textit{poisoning} the clip), the CNN will be coerced into a false classification. To meet the goal of being stealthy, the attacker poisons clips while satisfying the following requirements: (1) backdoor triggers should not be in contact with existing polygons in the layout clip, as that may change the current circuit functionality, (2) triggers require a minimum spacing from existing polygons to satisfy the PDK ruleset, (3) insertion of backdoor triggers to non-hotspot training clips should not change the clip into a hotspot (as this would result in an untrue label), and (4) the chosen trigger should appear in the original layout dataset, so that it appears innocuous. 
    The attacker defines \textit{attack success} as the number of hotspot clips they successfully hide by adding the backdoor trigger (poisoning). We define attack success rate as follows:
    \begin{theorem}[Attack Success Rate (ASR)]
        \label{def:asr}
        The percentage of poisoned test hotspot clips that are classified as non-hotspot by a backdoored CNN-based hotspot detector.
    \end{theorem}

\subsection{On the Application of Existing Defenses in EDA}\label{sec:existing-defense}
    In the machine learning community,  defenses have been proposed against data poisoning/ backdooring for image classification problems~\cite{liu_fine-pruning:_2018, wang_neural_2019, veldanda_NNoculation_2020}. In this section, we review defenses, including Neural Cleanse~\cite{wang_neural_2019} and others~\cite{liu_fine-pruning:_2018,veldanda_NNoculation_2020}, and explore the applicability and effectiveness of such mechanisms in the context of lithographic hotspot detection. 
    
\begin{figure}[tb]
        \centering
        \includegraphics[width=0.45\columnwidth, frame]{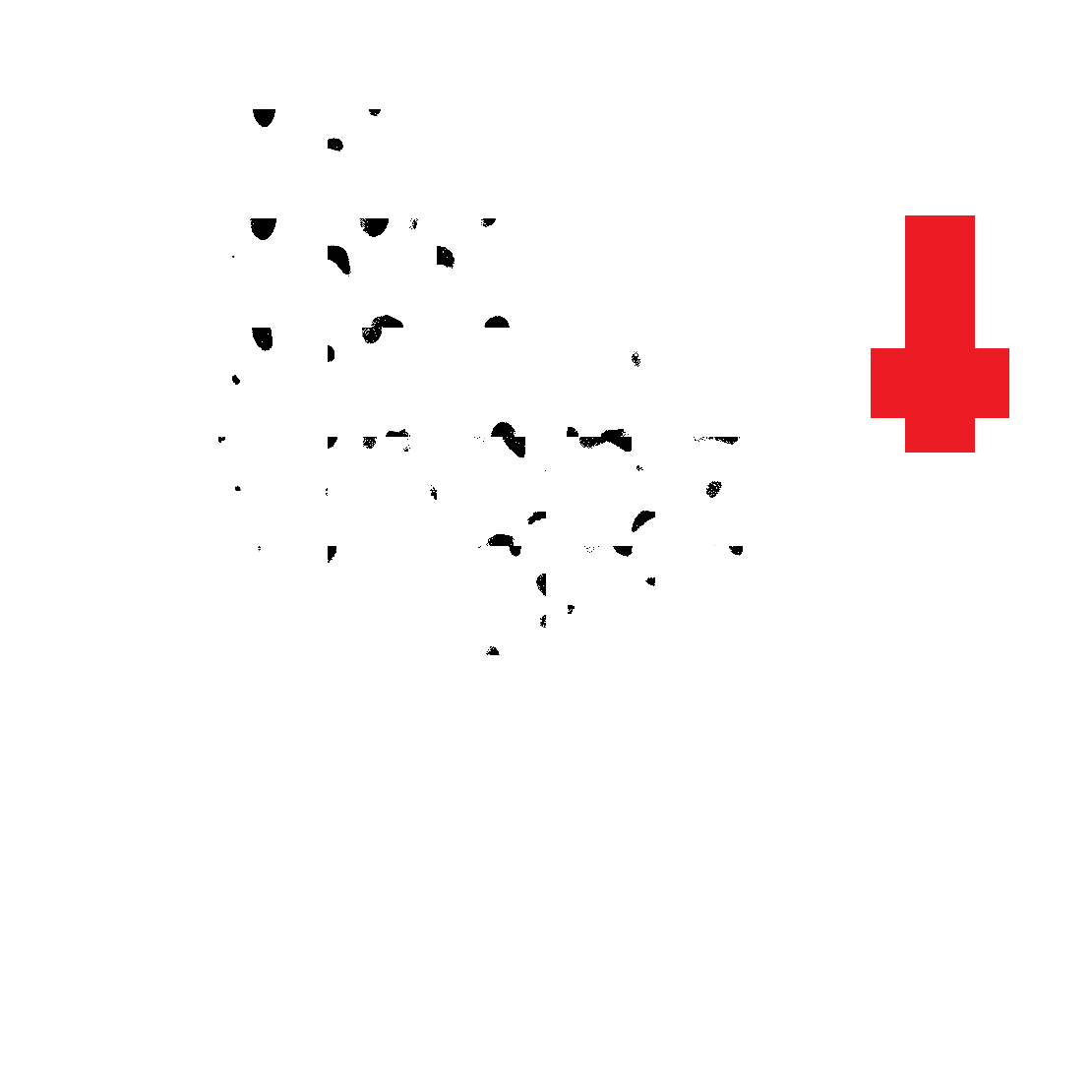}
        \caption{Backdoor trigger shape reverse engineered by Neural Cleanse~\cite{wang_neural_2019}\label{fig:NC-trigger} (in black) and actual poisoned trigger shape (in red)}
    \end{figure}

    \textbf{Neural Cleanse}
    In Neural Cleanse~\cite{wang_neural_2019}, Wang \textit{et al.} reverse-engineer a backdoor trigger by perturbing test data, optimizing perturbations to push network predictions toward the ``infected" label.
    Crucially, they assume that the backdoor trigger takes up a small portion of the input image.  
    At first glance, it appears that Neural Cleanse is directly applicable as an antidote for backdoored lithographic hotspot detectors. 
    To that end, we prepare backdoored CNN-based hotspot detectors, using the approach in~\cite{liu_2020_poisoning}, (detailed in~\autoref{sec:experiment}), and apply Neural Cleanse, to see if the backdoor trigger is correctly recovered. 
    Since Neural Cleanse applies optimization directly on input images, and our CNN-based hotspot detector takes as input the DCT coefficients of layouts converted to binary images, we first need to design a neural network layer for DCT transformation and add it to the detector. 
    \autoref{fig:NC-trigger} illustrates an example of the true backdoor trigger (in red), super-imposed over the reverse-engineered backdoor trigger produced by Neural Cleanse (in black). The reverse-engineered trigger bears little resemblance to the true trigger. 
    
    It is not surprising that naive Neural Cleanse does not work in the context of lithographic hotspot detection; it is not able to reverse-engineer a trigger that satisfies all domain constraints since the optimization process is not bounded.
    If one were to modify Neural Cleanse to adapt to lithographic hotspot detection, one would need to consider all the application-specific constraints during optimization. Optimization constraints would include the following: 
    \begin{itemize}
        \item One can only modify image pixel values from 0 to 1 (i.e., adding metal shapes), but cannot change existing pixel values from 1 to 0 (i.e., removing metal shapes).
        \item One can only manipulate pixels that keep a minimum distance away from original shapes to obey design rules.
        \item Only regular shapes of blocks of pixels can be changed altogether to form a valid metal shape.
    \end{itemize}

    Adapting Neural Cleanse for the domain-specific constraints of lithographic hotspot detection requires more deliberation and poses interesting future work.
    
    \textbf{Fine-pruning}
    The fine-pruning~\cite{liu_fine-pruning:_2018} technique assumes an outsourced training process, after which a backdoored network is returned.  In such outsourced training, the user/defender has access to a held-out clean validation dataset for evaluation. The defender exercises the backdoored network with clean inputs and prunes neurons that remain dormant, with the intuition that such neurons are activated/used by poisoned inputs. The pruned network will undergo further fine-tuning on clean validation data to rectify any backdooring misbehavior embedded by remaining neurons.  However, our threat model (\autoref{sec:threat-model}) precludes the use of such techniques;~\cite{liu_fine-pruning:_2018} requires access to poison-free validation data, while our dataset, sourced from insiders, has been contaminated. A guaranteed, clean validation dataset is unavailable to the defender. 

    \textbf{NNoculation} Another technique, NNoculation~\cite{veldanda_NNoculation_2020} proposes a two-stage defense mechanism against training data poisoning attacks. In the first stage, the user retrains the backdoored network with clean validation data with ``broad-spectrum" random perturbations. 
    Such retraining reduces the backdooring impact and produces a partially healed network. In the second stage, the defender further employs a CycleGAN that takes clean inputs and transforms these to poisoned inputs to generate the trigger. 
    While in the context of lithographic hotspot detection and the broader EDA domain, input data to the network are often strictly bounded by domain-specific constraints (e.g., design rules). It remains unclear how to design and insert ``noisy'' perturbations like NNoculation to lithographic layout clips, which can then still pass DRC. Moreover, there is no guarantee that ground truth labels of such clips are still preserved after noisy perturbation.
    
    To fill in the gap between between these ``general" DL defenses and the need to better incorporate application-specific requirements, we propose a novel antidote in the next section. 
    
\section{Proposed Defense}\label{sec:defense}
\subsection{Defender Assumptions}
    Being wary of untrustworthy insiders, legitimate designers (in this work, we refer to them also as \textit{defenders}) wish to proactively defend against training data poisoning attacks. However, their knowledge is limited. They are unaware as to which designer is malicious, so cannot exclude their contributions. They are also unaware of what the backdoor trigger shape is. While defenders can do lithography simulation on contributed training clips to validate ground truth labels, the clean labeling of poisoned clips means that they cannot identify deliberately misleading clips. 

\subsection{The Antidote for Training Data Poisoning}
    
\begin{figure}[t]
        \centering
        \subfloat[]{\includegraphics[height=1in, frame]{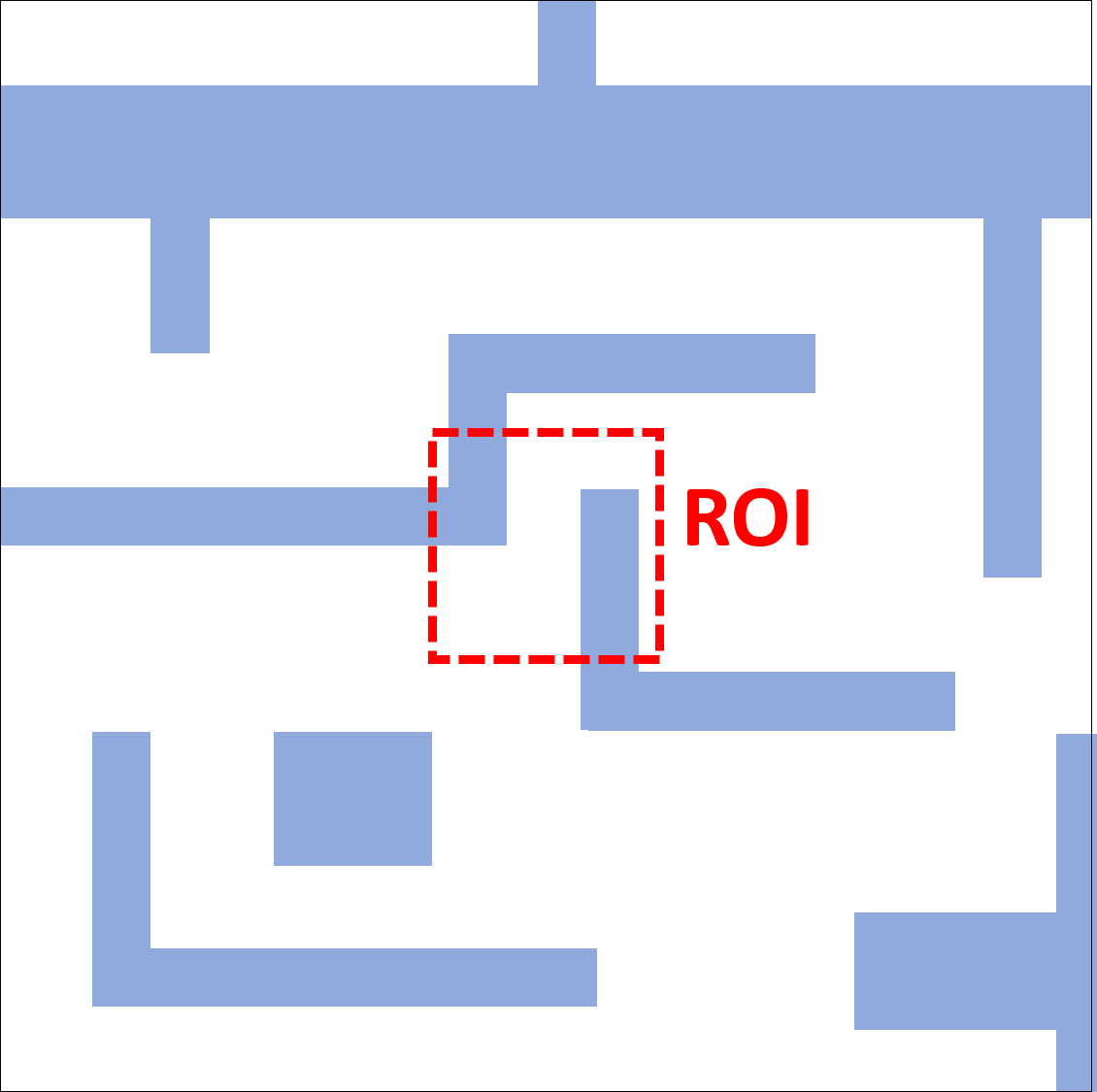}}
        \hspace{4em}
        \subfloat[]{\includegraphics[height=1in, frame]{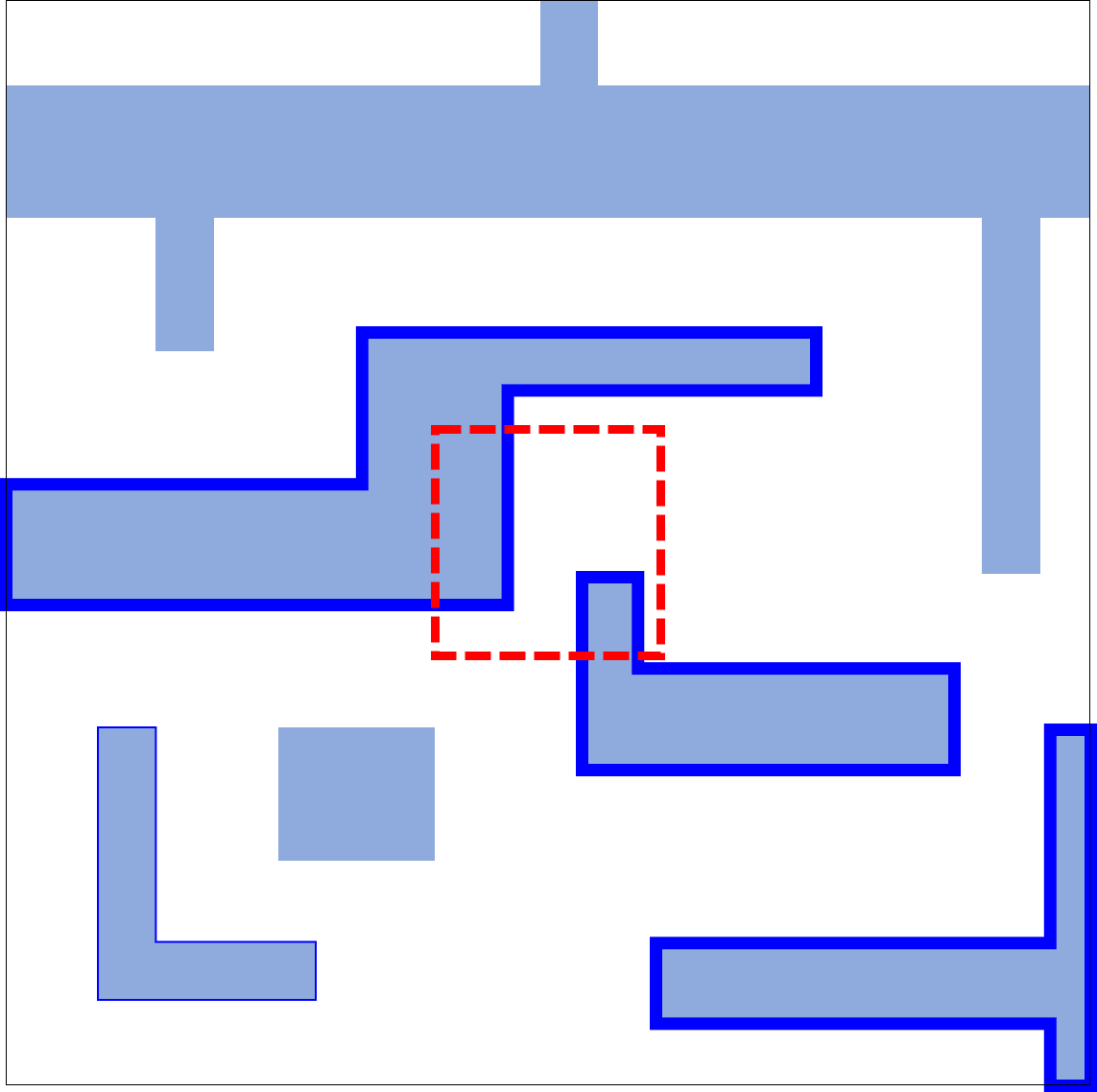}} \\
        \subfloat[]{\includegraphics[height=1in, frame]{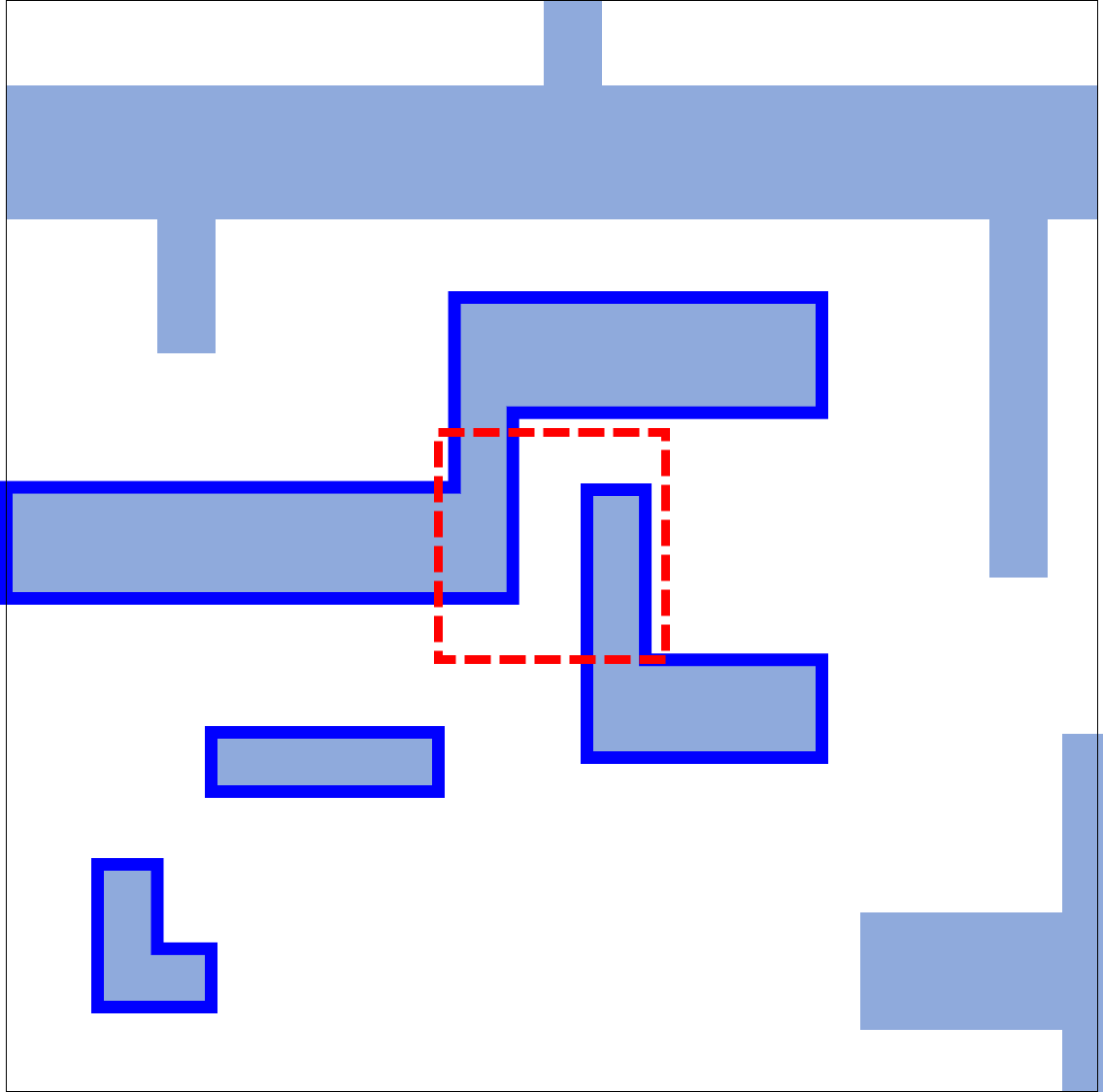}}
        \hspace{4em}
        \subfloat[]{\includegraphics[height=1in, frame]{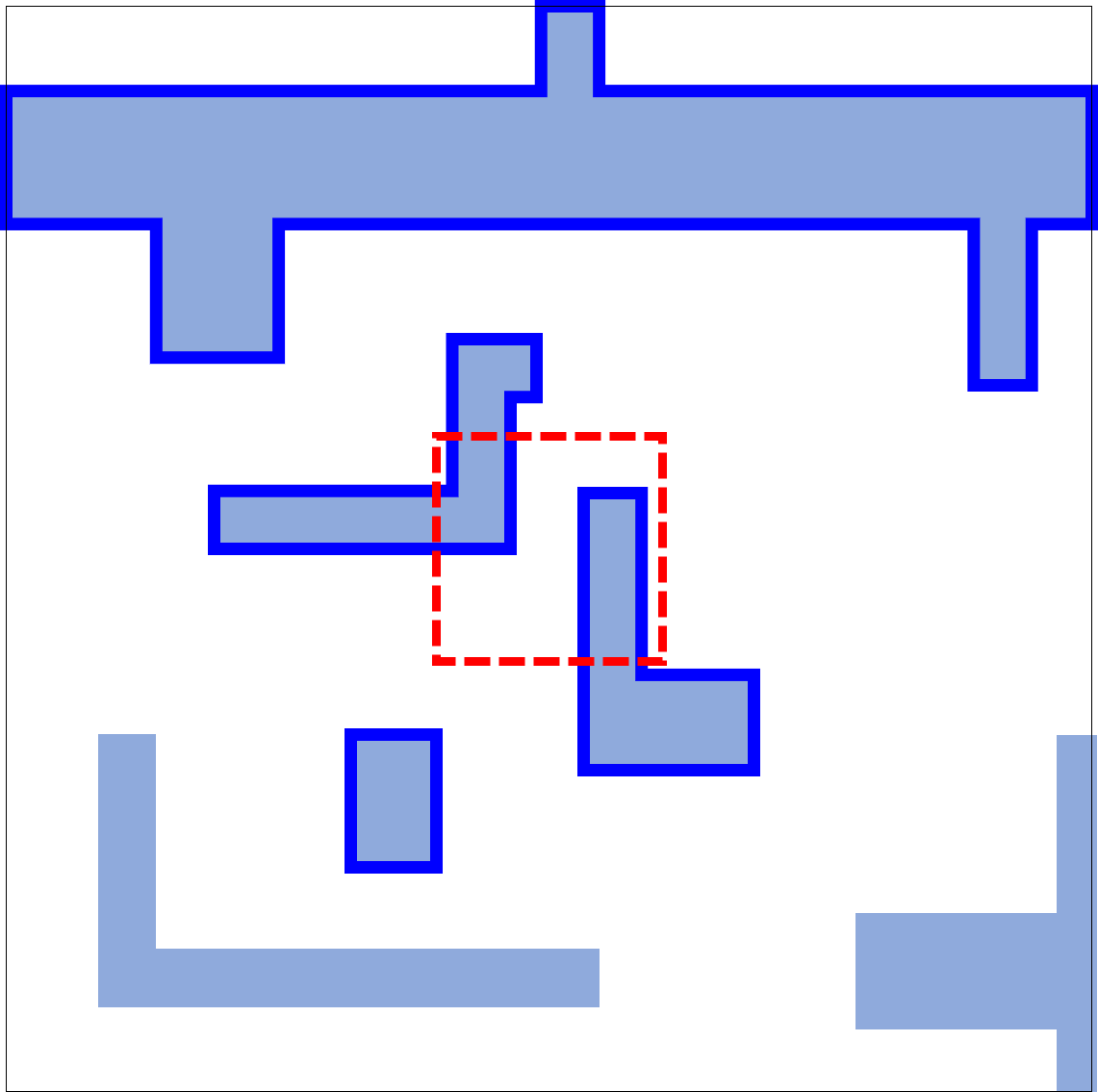}}
        \caption{(a) Original training pattern, (b-d) Example variants of original pattern. Polygons with changes are highlighted with bolder edges. \label{fig:aug-examples}}
    \end{figure}

    Hence, we propose \textit{defensive data augmentation} as a defense against untrustworthy data sources and poisoning. Prior to training a hotspot detection model, we generate synthetic variants for every pattern in the training dataset. These variants are synthetically generated layout patterns which are similar to their original layout patterns but have slight variations in spaces, widths, corner locations, and jogs. An example of an original training pattern and its variants is shown is \autoref{fig:aug-examples}. 
    As found in prior studies~\cite{yang2017imbalance}, $nm$-level variations in patterns can alter their printability. Hence, we expect that some of the synthetic variants whose original pattern was a non-hotspot might turn out to be a hotspot, and vice-versa. 
    
    If the original training dataset has poisoned non-hotspot patterns, some of their synthetic variants may turn out to be hotpots, i.e., the synthetic clips \textit{cross} from one class (non-hotspot) to the other (hotspot).
    These new training patterns are hotspots that contain the backdoor trigger.  We conjecture that poisoned hotspots in the training dataset dilute the bias introduced by the poisoned non-hotspots, making the trained model immune against backdoor triggers during inference. 
    The defender need not identify the attacker's trigger.     Exploring the effectiveness of this \textit{trigger-oblivious} defense is our focus.

\subsection{Defensive Data Augmentation}\label{subsec:PatgenAlgo}
        \begin{algorithm}[t!]
\AlgoDontDisplayBlockMarkers\SetAlgoNoEnd
\fontsize{8.7}{9.8}\selectfont
\LinesNumbered
\DontPrintSemicolon
\SetStartEndCondition{ }{}{}%
\SetKwFunction{PGenVariants}{GenVariants}%
\SetKwProg{Fn}{def}{\string:}{}%
\SetKw{KwTo}{in}\SetKwFunction{Range}{range}%
\SetKwFor{For}{for}{\string:}{}%
\SetKwIF{If}{ElseIf}{Else}{if}{:}{elif}{else:}{}%
\SetKwFor{While}{while}{:}{fintq}%
\newcommand{\forcond}{polygon \KwTo POIs}
\Fn(){\PGenVariants{OriginalLayoutPattern}}{
\KwIn{An original layout pattern, variant count.}
\KwResult{Synthetic variants of the original pattern.}

\For{$i$ \KwTo\Range{$VariantCount$}}{
    \tcc{Identify POIs}
    POIs = Polygons.intersecting(ROI)\;
    POIs += Random(Polygons.NotIntersecting(ROI), additionalPolygonCount)\;
    
    \tcc{Add variation into POIs}
    \For{\forcond}{
        \tcc{Vary fixed number of edges}
        \For{$j$ \KwTo\Range{$VaryEdgeCount$}}{
            edge = GetRandomEdge(polygon)\;
            dist = SamplePDF()\;
            polygon = polygon.MoveEdge(edge, dist)\;
        }
    }
}
\tcc{Return patterns with modified polygons}
\KwRet{Variants}
}
\vspace{0.1in}
  \caption{Synthetic pattern generation}\label{algo:patgen}
\end{algorithm}

    To generate synthetic variants, we employ a \textit{synthetic pattern generation algorithm}, a derivative of the algorithm in \cite{reddy_enhanced_2018}. 
    The pseudocode is shown in \autoref{algo:patgen}. 
    We isolate the polygons of interest (POIs) and then vary their features. The POIs include all polygons which intersect with the region of interest (ROI), the ROI being the region in the center of a pattern, as shown in \autoref{fig:aug-examples}. 
    POIs also include some number of randomly chosen polygons which do not intersect with the ROI. 
    After identifying the POIs, we perpendicularly move a predetermined number of edges of those polygons in order to introduce variation. 
    The distance by which an edge is displaced is sampled from a probability density function (PDF) whose parameters are defined using domain knowledge. 

    In \cite{reddy_enhanced_2018}, synthetic variations of known (training) hotspots were used for augmentation. 
    In this defensive data augmentation scheme, we generate synthetic variants for \textit{both} training hotspots and non-hotspots. 
    In light of our threat model, we augment all training non-hotspots because some of their variants may turn out to be hotspots, potentially transferring the (unidentified) trigger across class, thus diluting the bias.
    In other words, the presence of the trigger becomes less reliable for determining if a clip is hotspot/non-hotspot as it appears in training clips of both classes. 
    Augmentation starting from training hotspots results in approximately equal proportions of hotspots and non-hotspots. Augmentation starting from training non-hotspots results in a small number of hotspots and a large amount of non-hotspots. Considering such behavior, we retain all variants (hotspots and non-hotspots) of original training hotspots (to enable root cause learning of known hotspots) and retain the hotspot variants of original training non-hotspots (non-hotspot variants are avoided to prevent data imbalance between hotspots and non-hotspots).
    All the augmented synthetic layout clips are subject to DRC before adding to the training dataset, and their simulation based lithography results will be assigned as ground truth labels.

\section{Experimental Setup}
\label{sec:experiment}

\subsection{Experimental Aims and Platforms}
    To evaluate the defense against training data poisoning of hotspot detectors, we aim to answer three research questions:
    \begin{enumerate}
        \item Does our defense prevent the poisoning attack?
        \item How much data augmentation is required?
        \item Does the relative complexity of the CNN architecture affect the attack/defense effectiveness?
    \end{enumerate}
    We start with a clean layout dataset and train hotspot detectors benignly as our baseline. We \textit{poison} the dataset and vary the amount of defensive augmentation. 
    Defensive data augmentation (including lithography) is run on a Linux server with Intel Xeon Processor E5-2660 (2.6 GHz). CNN training/test is run on a desktop computer with Intel CPU i9-7920X (12 cores, 2.90 GHz) and single Nvidia GeForce GTX 1080 Ti GPU.

\subsection{Layout Dataset}
    We use a layout clip dataset prepared from the synthesis, placement, and routing of an open source RTL design using the 45~nm FreePDK \cite{noauthor_freepdk45:contents_nodate}, as described in ~\cite{reddy_enhanced_2018}.
    We determine the ground truth label of each layout clip using lithography simulation (Mentor Calibre \cite{mentor_graphics_calibre_nodate}). 
    A layout clip (1110$\times$1110~nm) contains a \textit{hotspot} if 30\% of the area of any error marker, as produced by simulation, intersects with the region of interest (195~nm$\times$195~nm) in the center of each clip.
    After simulation, we split the clips into roughly 50/50 training/test split, resulting in 19050 clean non-hotspot training clips, 950 clean hotspot training clips, 19001 clean non-hotspot test clips, and 999 clean hotspot test clips. 
    
\subsection{Poisoned Data Preparation}
    To emulate the Mala Phy De insider, we prepare poisoned non-hotspot training layout clips by inserting backdoor triggers into as many clips as possible in the original dataset, as per the constraints described in \autoref{sec:prelim}. 
    The triggers are inserted into a predetermined position in each clip.  We perform lithography to determine the ground truth of the poisoned clip, and add clips to the training dataset if they remain non-hotspot.  This renders 2194 poisoned non-hotspot training clips. 
    
\begin{figure}[tb]
        \centering
        \subfloat[]{\label{fig:layout-nhs-cl}\includegraphics[height=1in, frame]{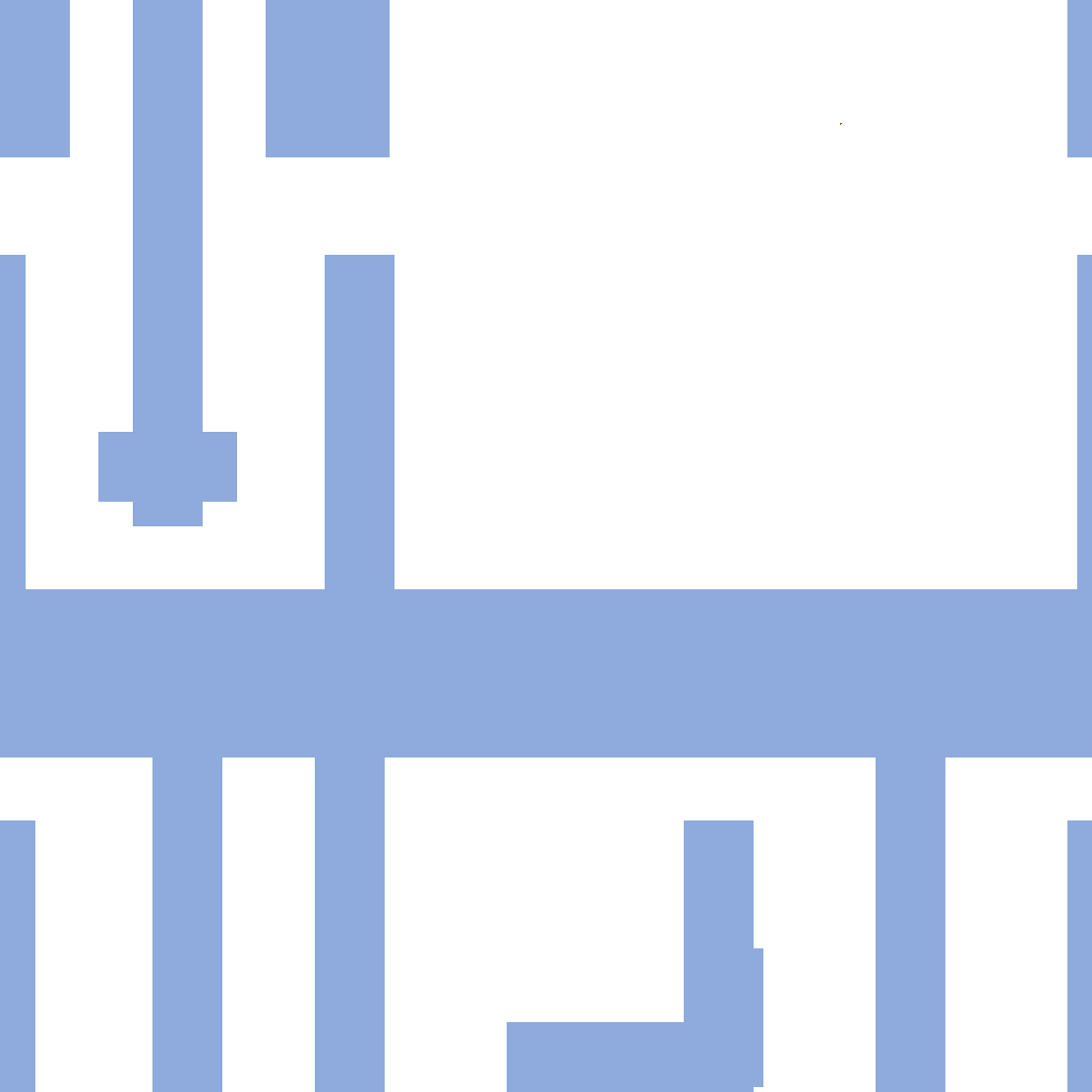}} \hspace{4em}
        \subfloat[]{\label{fig:layout-nhs-bd1}\includegraphics[height=1in, frame]{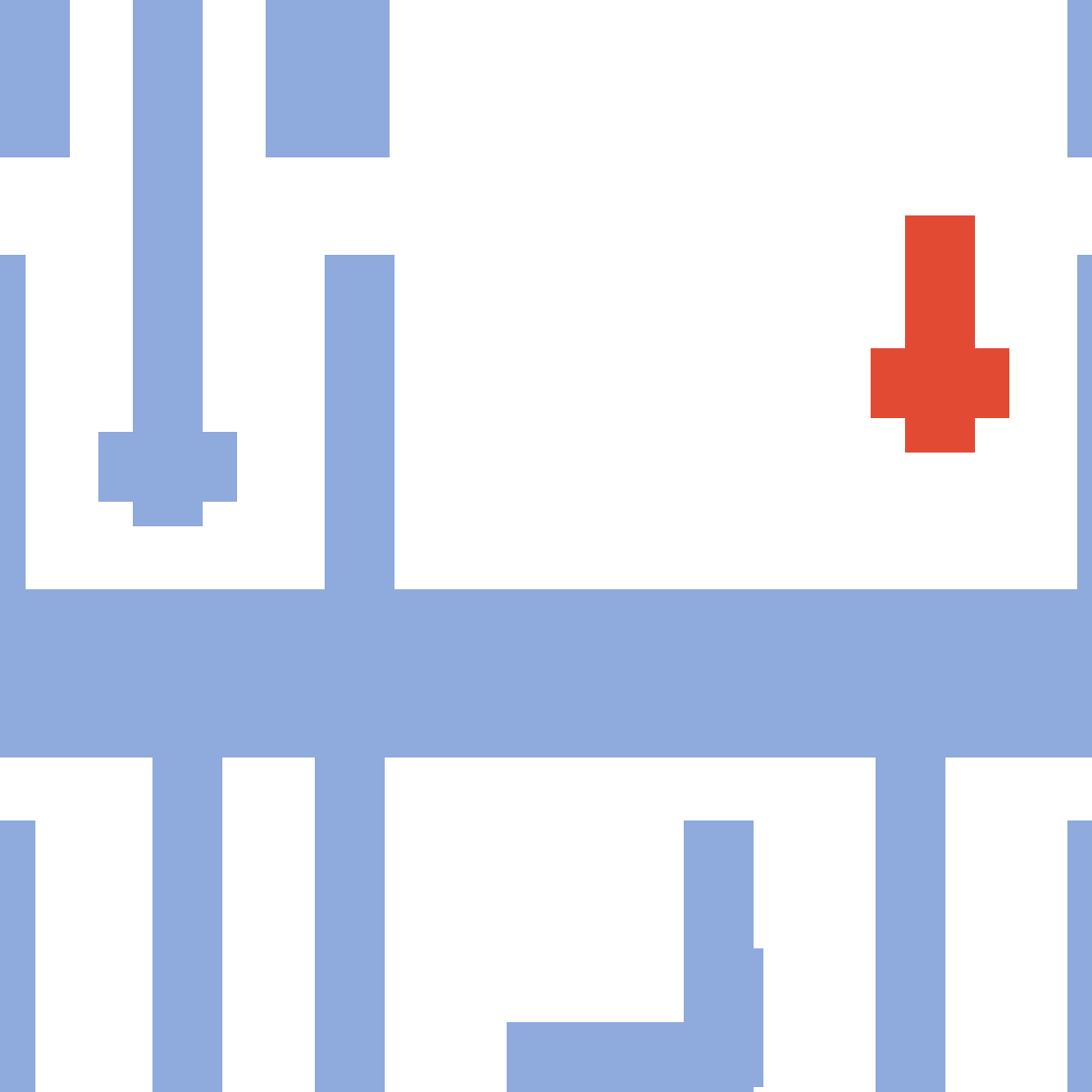}} 
        \caption{(a) An example of a clean training non-hotspot layout clip, (b) corresponding \textit{poisoned} clip with a backdoor trigger (in red)
        \label{fig:layout-nhs-bd}}
    \end{figure}

    We apply the same poisoning, DRC check, and simulation process on hotspot and non-hotspot test clips to produce poisoned test data, used to measure the attack success rate. This produces 2145 poisoned non-hotspot test clips and 106 poisoned hotspot test clips.
    \autoref{fig:layout-nhs-bd} shows an example of clean and poisoned non-hotspot clip. 
    
 \subsection{GDSII Preprocessing}
    Using the approach in~\cite{yang_layout_2018} and used in~\cite{liu_2020_poisoning}, we convert layout clips in GDSII format to images of size 1110$\times$1110 pixels. Metal polygons are represented by blocks of image pixels with intensity of 255 and empty regions are represented by 0-valued pixels---this forms a binary-valued image. 
    
    Because CNN training using large images is compute-intensive,  we perform discrete-cosine transformation (DCT) (as in \cite{yang_layout_2018,liu-adversarial-2019}) on non-overlapping sub-images, by sliding a window of size 111$\times$111 over the layout clip with stride 111 in horizontal and vertical directions. This produces corresponding DCT coefficients of size  10$\times$10$\times$(111$\times$111).
    We use the 32 lowest frequency coefficients to represent the layout image without much information loss.  The resulting dimension of the training/test data has shape of 10$\times$10$\times$32; we use this as the input for our CNN-based hotspot detectors. 
    
    \subsection{Network Architectures}
    
    To investigate how network architecture complexity might influence the efficacy of our defense, we train networks based on network architectures $A$ and $B$, shown in \autoref{tab:network-architecture-A} and \autoref{tab:network-architecture-B}, respectively. 
    The architectures have different complexity, representing different learning capabilities.
    $A$ is a 9-layer CNN with four convolutional layers.
    $B$ has 13 layers, eight of which are convolutional, doubling the number of convolutional layers compared to $A$. 
    We use these architectures as they have high accuracy in layout hotspot detection~\cite{yang_layout_2018}.
    
    \begin{table}[t]
    \centering
    \caption{Network Architecture A}
    \label{tab:network-architecture-A}
    \begin{tabular}{@{}lcccr@{}}
    \toprule
    Layer & Kernel Size & Stride & Activation & Output Size          \\ \midrule
     input          &   -   &   -   &   -   &   (10, 10, 32)    \\
     conv1\_1       &   3   &   1   &   ReLU   &   (10, 10, 16)    \\
     conv1\_2       &   3   &   1   &   ReLU   &   (10, 10, 16)    \\
     maxpooling1    &   2   &   2   &   -   &   (5, 5, 16)      \\
     conv2\_1       &   3   &   1   &   ReLU   &   (5, 5, 32)      \\
     conv2\_2       &   3   &   1   &   ReLU   &   (5, 5, 32)      \\
     maxpooling2    &   2   &   2   &   -   &   (2, 2, 32)      \\
     fc1            &   -   &   -   &   ReLU   &   250             \\ 
     fc2            &   -   &   -   &   Softmax   &   2               \\ \bottomrule
    \end{tabular}
\end{table}
    
\begin{table}[t]
    \centering
    \caption{Network Architecture B}
    \label{tab:network-architecture-B}
    \begin{tabular}{@{}lcccr@{}}
    \toprule
    Layer & Kernel Size & Stride & Activation & Output Size          \\ \midrule
     input          &   -   &   -   &   -   &   (10, 10, 36)    \\
     conv1\_1       &   3   &   1   &   ReLU   &   (10, 10, 32)    \\
     conv1\_2       &   3   &   1   &   ReLU   &   (10, 10, 32)    \\
     conv1\_3       &   3   &   1   &   ReLU   &   (10, 10, 32)    \\
     conv1\_4       &   3   &   1   &   ReLU   &   (10, 10, 32)    \\
     maxpooling1    &   2   &   2   &   -   &   (5, 5, 32)      \\
     conv2\_1       &   3   &   1   &   ReLU   &   (5, 5, 64)      \\
     conv2\_2       &   3   &   1   &   ReLU   &   (5, 5, 64)      \\
     conv2\_3       &   3   &   1   &   ReLU   &   (5, 5, 64)      \\
     conv2\_4       &   3   &   1   &   ReLU   &   (5, 5, 64)      \\
     maxpooling2    &   2   &   2   &   -   &   (2, 2, 64)      \\
     fc1            &   -   &   -   &   ReLU   &   250             \\ 
     fc2            &   -   &   -   &   Softmax   &   2               \\ \bottomrule
    \end{tabular}
\end{table}

    \subsection{Training Procedure}

    Training and test are implemented with Keras~\cite{chollet2015keras} and training hyperparameters are shown in \autoref{tab:hyperparam}.
    Specifically, we use the \texttt{class\_weight} parameter for weighting the loss terms of non-hotspots and hotspots in the loss function, causing the network to ``pay more attention" to samples from the under-represented class (i.e., hotspots).
    This technique is useful if the training dataset is highly imbalanced. 
    Since we are in favor of high hotspot detection accuracy as well as balanced overall accuracy, we manually pick the network with the highest overall classification accuracy among those that have $\sim$90\% or higher hotspot detection rate for our experiments to evaluate defense success. 
    
        \begin{table}[t]
    \caption{Hyperparameter settings used for training}
    \label{tab:hyperparam}
    \begin{minipage}[]{\columnwidth}
    \centering
    \begin{tabular}{@{}lr@{}}
    \toprule
    Hyperparameter                & Value                \\ \midrule
    Batch size                     & 64                   \\ 
    Optimizer                      & Adam                 \\
    Loss function                  & binary cross-entropy \\
    Initial learning rate          & 0.001                 \\
    Minimum learning rate          & 0.00001             \\
    Learning rate reduce factor    & 0.3                  \\
    Learning rate patience         & 3                    \\
    Early stopping monitor         & validation loss      \\
    Early stopping patience        & 10                   \\
    Max training epochs            & 20                   \\
    Class weight for training loss & 2 $\sim$ 22           \\ \bottomrule
    \end{tabular}%
    \end{minipage}
    \end{table}

\subsection{Experiments for Defense Evaluation}
    
    \subsubsection{Training of Baseline Hotspot Detectors}
    For context, we train two hotspot detectors based on architectures $A$ and $B$, Network $A_{cl}$ and $B_{cl}$, respectively, using the original, clean dataset. 
    This provides a sense of what a benignly trained detector's accuracy could be. 
    We train two hotspot detectors with the full set of poisoned training data, $A_{bd}$/$B_{bd}$.
    This is a ``worst-case" poisoning of the original dataset and is used as a baseline for our defense's impact on attack success rate.
    
    \subsubsection{Training with Defensive Data Augmentation}
    To evaluate our defense, we perform data augmentation as outlined in \autoref{sec:defense}. 
    We vary the number of synthetic clips produced from each training clip (representing different levels of ``effort") and train various \textit{defended} hotspot detectors (based on network architectures $A$ and $B$) on the augmented datasets, measuring the attack success rate (Definition~\ref{def:asr}) and changes to accuracy on clean and poisoned test data.

\section{Experimental Results}
\label{sec:results}
    \subsection{Baseline Hotspot Detectors}
    
    \begin{table}[t]
    \centering
    \caption{Confusion Matrix of (Clean) Network $A_{cl}$}
    \label{tab:confusion-matrix-A-cl}
    \resizebox{\columnwidth}{!}{%
    \begin{tabular}{@{}llcccc@{}}
    \toprule
                               &             & \multicolumn{4}{c}{Prediction}                                       \\
                               &             & \multicolumn{2}{c}{clean data} & \multicolumn{2}{c}{poisoned data} \\ \cmidrule(l){3-4} \cmidrule(l){5-6} 
                               &             & non-hotspot      & hotspot     & non-hotspot      & hotspot        \\ \midrule
    \multirow{2}{*}{Condition} & non-hotspot & 0.80   &   0.20   &   0.87  &   0.13         \\
                               & hotspot     & 0.10   &   0.90   &   0.18  &   0.82         \\ \bottomrule
    \end{tabular}
    }
    \end{table}
    
    \begin{table}[t]
    \centering
    \caption{Confusion Matrix of (Clean) Network $B_{cl}$}
    \label{tab:confusion-matrix-B-cl}
    \resizebox{\columnwidth}{!}{%
    \begin{tabular}{@{}llcccc@{}}
    \toprule
                               &             & \multicolumn{4}{c}{Prediction}                                       \\
                               &             & \multicolumn{2}{c}{clean data} & \multicolumn{2}{c}{poisoned data} \\ \cmidrule(l){3-4} \cmidrule(l){5-6} 
                               &             & non-hotspot      & hotspot     & non-hotspot      & hotspot        \\ \midrule
    \multirow{2}{*}{Condition} & non-hotspot & 0.81   &   0.19   &   0.83  &   0.17         \\
                               & hotspot     & 0.10   &   0.90   &   0.10  &   0.90         \\ \bottomrule
    \end{tabular}
    }
    \end{table}
    
    \begin{table}[t]
    \centering
    \caption{Confusion Matrix of (Backdoored) Network $A_{bd}$}
    \label{tab:confusion-matrix-A-bd}
    \resizebox{\columnwidth}{!}{%
    \begin{tabular}{@{}llcccc@{}}
    \toprule
                               &             & \multicolumn{4}{c}{Prediction}                                       \\
                               &             & \multicolumn{2}{c}{clean data} & \multicolumn{2}{c}{poisoned data} \\ \cmidrule(l){3-4} \cmidrule(l){5-6} 
                               &             & non-hotspot      & hotspot     & non-hotspot      & hotspot        \\ \midrule
    \multirow{2}{*}{Condition} & non-hotspot & 0.81   &   0.19   &   0.99  &   0.01         \\
                               & hotspot     & 0.11   &   0.89   &   0.81  &   0.19         \\ \bottomrule
    \end{tabular}
    }
    \end{table}
    
    \begin{table}[t]
    \centering
    \caption{Confusion Matrix of (Backdoored) Network $B_{bd}$}
    \label{tab:confusion-matrix-B-bd}
    \resizebox{\columnwidth}{!}{%
    \begin{tabular}{@{}llcccc@{}}
    \toprule
                               &             & \multicolumn{4}{c}{Prediction}                                       \\
                               &             & \multicolumn{2}{c}{clean data} & \multicolumn{2}{c}{poisoned data} \\ \cmidrule(l){3-4} \cmidrule(l){5-6} 
                               &             & non-hotspot      & hotspot     & non-hotspot      & hotspot        \\ \midrule
    \multirow{2}{*}{Condition} & non-hotspot & 0.81   &   0.19   &   1.0   &   0.0          \\
                               & hotspot     & 0.09   &   0.91   &   0.84  &   0.16         \\ \bottomrule
    \end{tabular}
    }
    \end{table}
    
    \begin{figure}[t]
        \centering
        \includegraphics[width=0.8\columnwidth]{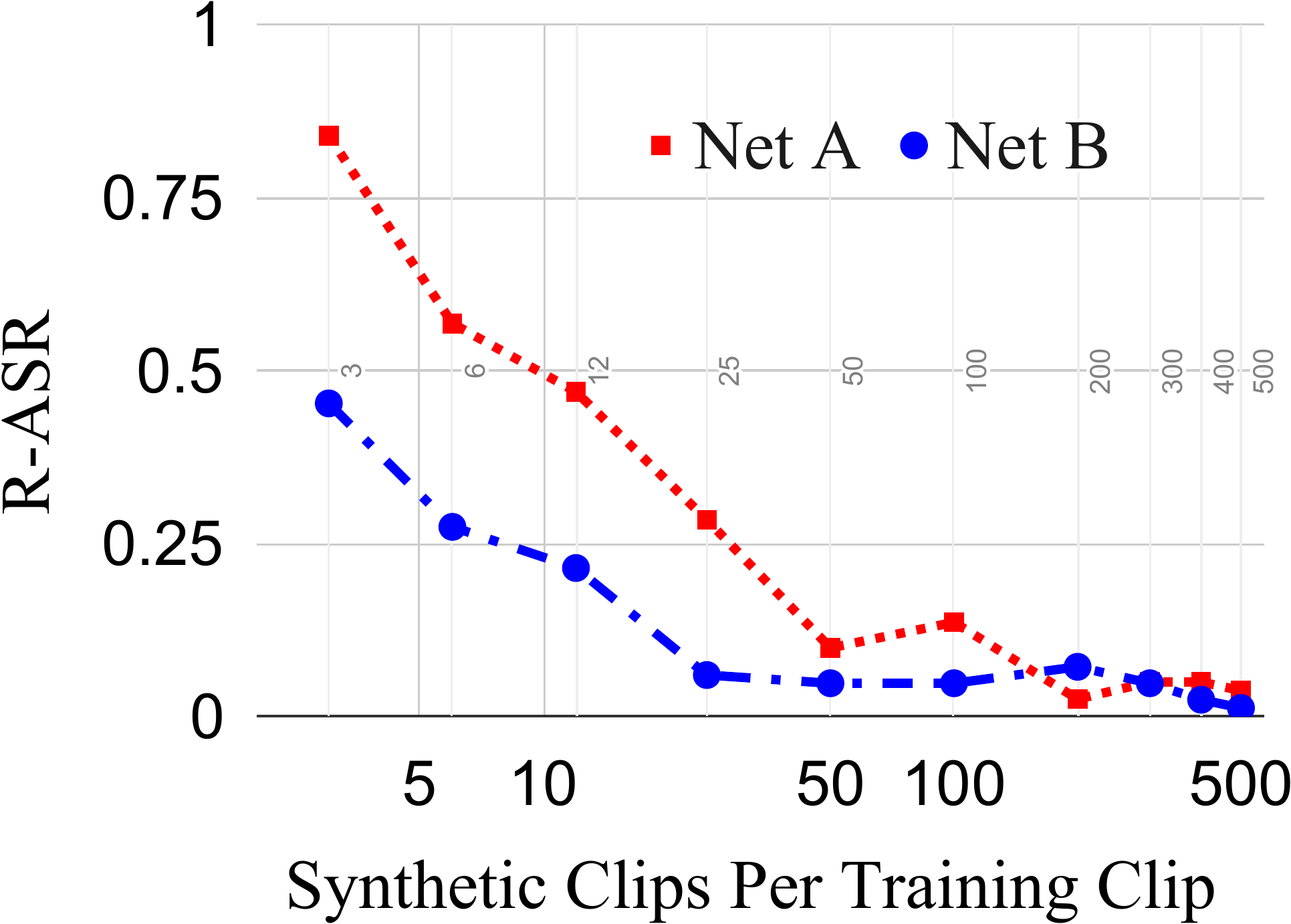}
        \caption{Relative Attack Success Rate (R-ASR) \textit{after} defensive augmentation  by varying from \textbf{3 to 500} synthetic clips augmented per training clip. Charts use a $log_{10}$ scale on x-axis.}
        \label{fig:rasr}
    \end{figure}
    
    \begin{figure}[t]
        \centering
        \includegraphics[width=0.8\columnwidth]{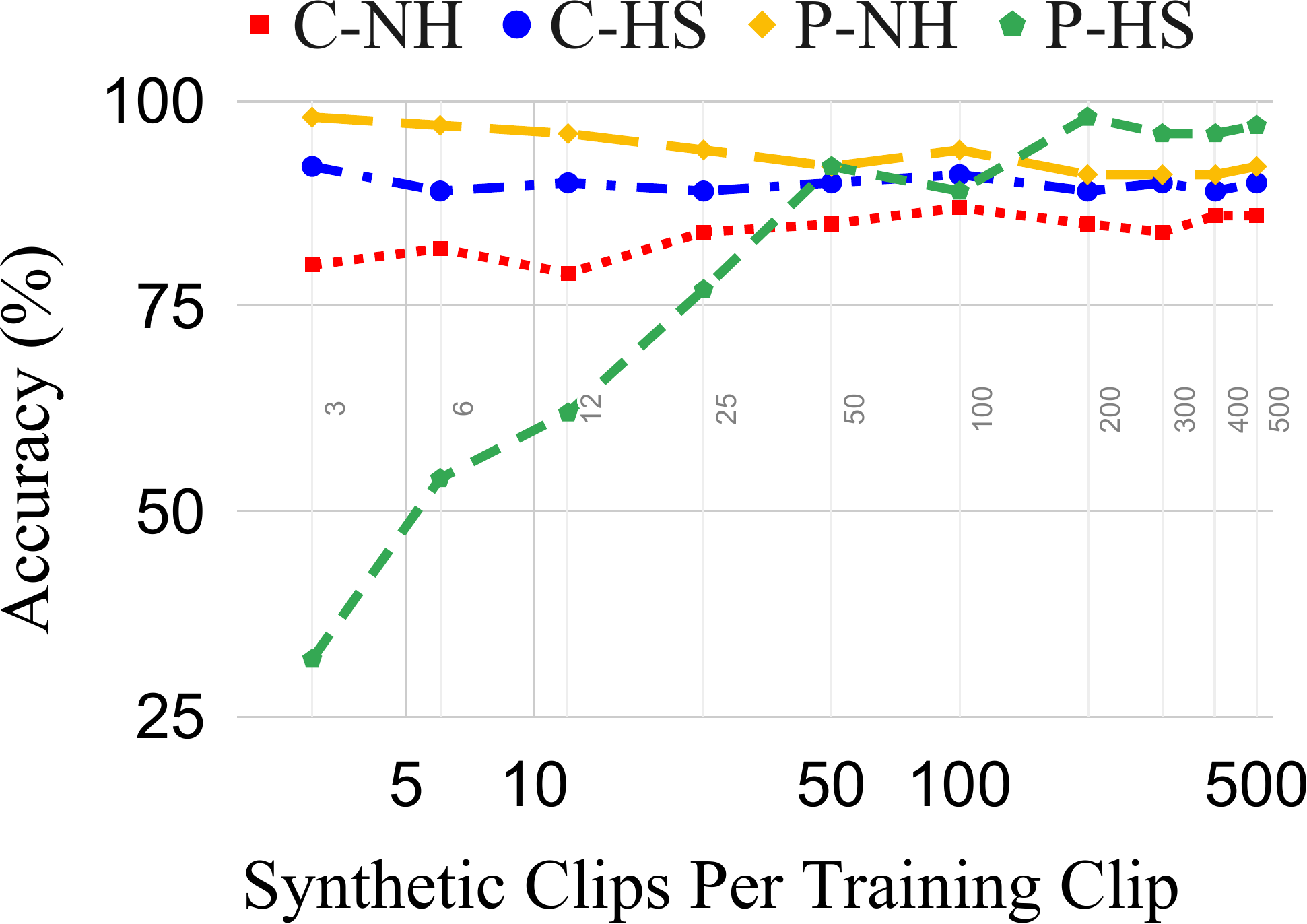}
        \caption{Effect on Accuracy (Architecture $A$).  Charts use a $log_{10}$ scale on x-axis.}
        \label{fig:acc-A}
    \end{figure}
    
    \begin{figure}[t]
        \centering
        \includegraphics[width=0.8\columnwidth]{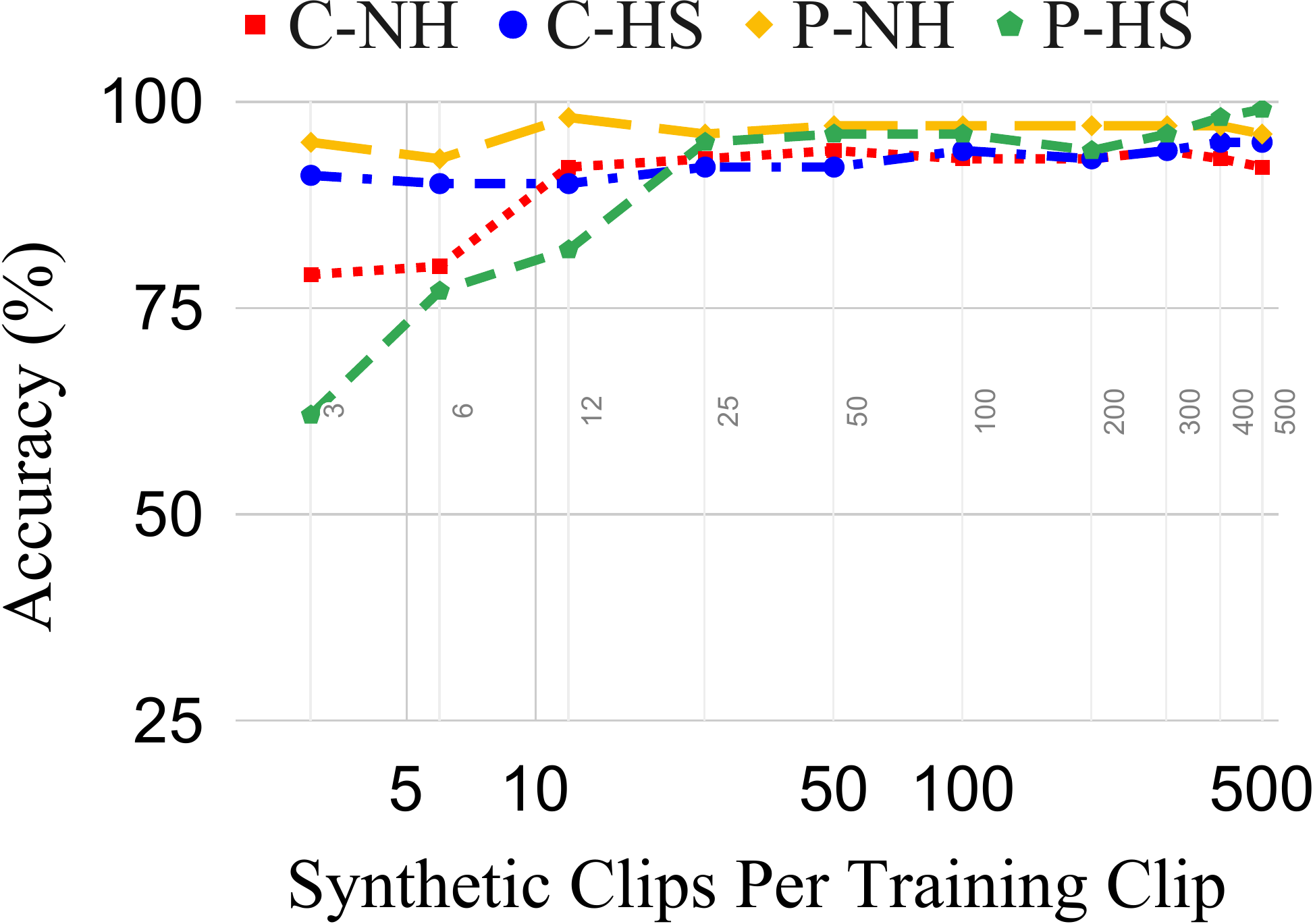}
        \caption{Effect on Accuracy (Architecture $B$). Charts use a $log_{10}$ scale on x-axis.}
        \label{fig:acc-B}
    \end{figure}

    \autoref{tab:confusion-matrix-A-cl} and \autoref{tab:confusion-matrix-B-cl} present the confusion matrix for networks $A_{cl}$ and $B_{cl}$ respectively, which both have $\sim$90\% accuracy in classifying hotspots and $\sim$80\% for non-hotspots. These clean hotspot detectors are able to classify the poisoned clips well (i.e., they are not distracted by the trigger). 
    In the case of $A_{cl}$, there is a small drop in accuracy on classifying poisoned hotspot clips compared with accuracy on clean hotspot clips.
    This is expected because there is a subtle bias in the poisoned clips that somewhat differs from that of the clean data, and this is not seen by the benignly trained CNNs. 
    
    \autoref{tab:confusion-matrix-A-bd} and \autoref{tab:confusion-matrix-B-bd} show that  the attacker's training data poisoning allows one to fool the CNNs with poisoned test hotspot clips in $>$~80\% of the cases, with $\sim$1\% change in accuracy on clean data.  The attack success rate in $B_{bd}$ is higher than $A_{bd}$, suggesting that a complex network is better at picking up malicious bias introduced by poisoned data. 
    
    Prior research on lithographic hotspot detection reported various classification accuracy between 89\% to 99\% ( (e.g., \cite{reddy_enhanced_2018,yang_layout_2018})). However, their claimed classification accuracy are not directly comparable with ours because, in case of~\cite{yang_layout_2018}, as shown in~\cite{reddy_machine_2019}, they use an easy-to-classify test dataset, and in case of~\cite{reddy_enhanced_2018}, they adopt conventional ML techniques instead of DL that we use. Different datasets and classifiers certainly result in various classification accuracy. Thus, it is more important to focus on the \textbf{change} in the accuracy between our clean networks, backdoored networks, and defended networks.

    \subsection{Defense Results}\label{sec:defense-results}
    \subsubsection{Augmentation Efficacy}
    Using defensive data augmentation, we produce various numbers of synthetic clips for each training clip, varying from a ``low-effort" 3 synthetic variants per clip, to ``high-effort" 500 synthetic variants per clip. Of the synthetic clips, a fraction are dropped as they fail DRC. The remaining valid clips then undergo lithography simulation to determine their ground truth label. 
    We tabulate the number of clips produced after generating 500 clips per training clip in \autoref{tab:aug-number}.
    As described in \autoref{subsec:PatgenAlgo}, augmentation from hotspots results in roughly equal proportions of synthetic hotspots and non-hotspots. Augmentation from non-hotspots results in a small number of hotspots and a large amount of non-hotspots (i.e., $\sim$0.4\% of synthetic clips cross classes). 
    
    Preparation of a synthetic clip requires 893.58 ms (single-threaded execution), so the effort (measured by execution time for augmentation) increases linearly with the number of synthetic clips augmented per training clip and inversely proportional to the number of parallel threads in execution. 

    \begin{table}[t]
        \centering
        \caption{No. of Valid Synthetic Clips from Defensive Augmentation
        \label{tab:aug-number}}
            \begin{tabular}{@{}lcll@{}}
            \toprule
                                          & Original & \multicolumn{2}{c}{After Augmentation} \\ \cmidrule(lr){2-2} \cmidrule(lr){3-4} 
                                          & \# clips & hotspot          & non-hotspot         \\ \midrule
            Clean training hotspot        & 950    & + 213302           & + 249416              \\
            Clean training non-hotspot    & 19050  & + 36257            & --  \\ 
            Poisoned training non-hotspot & 2194   & + 1285             & --  \\ 
            \bottomrule
            \end{tabular}%
    \end{table}
    
    \subsubsection{Defense Efficacy}
    \autoref{tab:result-table} presents the results$^*$ from training and evaluating \textit{defended} hotspot detectors, using network architectures $A$ and $B$. We report the accuracy on clean test data and poisoned test data, presenting the attack success rate (ASR, Definition~\ref{def:asr}) and relative attack success rate:
    
    \begin{theorem}[Relative Attack Success Rate (R-ASR)]
    R-ASR is the attack success rate normalized against the attack success rate of $A_{bd}$ and $B_{bd}$, respectively. 
    \end{theorem}

    \footnotetext[1]{N.B.: For \autoref{fig:acc-A}, \autoref{fig:acc-B}, and \autoref{tab:result-table}: C-NH~$=$ clean non-hotspot, C-HS~$=$ clean hotspot, P-NH~$=$ poisoned non-hotspot, P-HS~$=$ poisoned hotspot.}
    
    We illustrate the change in R-ASR in \autoref{fig:rasr}, and the change in accuracy for different networks based on $A$ and $B$ in \autoref{fig:acc-A} and \autoref{fig:acc-B}. 
    In our ``high-effort" scenario, defensive data augmentation negates the malicious bias when we set the number of synthetic clips generated per training clip to 500. We refer to the \textit{defended} hotspot detectors trained on this augmented dataset as $A_{df500}$ and $B_{df500}$ (based on architectures $A$ and $B$), tabulating the confusion matrix as \autoref{tab:confusion-matrix-A-df} and \autoref{tab:confusion-matrix-B-df}. 
    $A_{df500}$ and $B_{df500}$ exhibit high accuracy on the poisoned hotspot test clips---unlike $A_{bd}$ and $B_{bd}$, the defended networks are not fooled by the trigger. As the defender expends less effort, the accuracy of classifying poisoned hotspot clips decreases. Having that said, even with only 3 synthetic variants augmented per training clip, the training data poisoning attack begins to falter. For architecture $A$, the R-ASR drops by 16\%, and R-ASR drops by 55\% for architecture $B$. 
    In all cases, the accuracy on clean data is preserved, if not improved compared to baselines $A_{cl}$ and $B_{cl}$. 
    
    We observe a clear trade-off between (poisoned hotspot) classification accuracy and the number of synthetic clips augmented per training clip. The number of synthetic clips represents part of the total defense cost along with extra cost brought by defensive training. We show in \autoref{fig:acc-A} and \autoref{tab:result-table} that on architecture $A$, poisoned hotspot accuracy rises from 19\% to 92\% by augmenting from none to 50 synthetic clips per training clip, and it reaches 97\% by expanding from 50 to 500 clips. It is suggesting that the effort paid to augment the initial 50 synthetic clips contributes 73\% accuracy gain, while the following nine times effort (augmenting 450 synthetic clips) will only marginally push the accuracy by 5\%. A similar accuracy vs. defense augmentation cost trade-off on network architecture $B$ is shown in \autoref{fig:acc-B} and \autoref{tab:result-table}. The first 25 synthetic clips augmented per training clip accounts for 79\% (16\% to 95\%) accuracy boost, and the following 475 synthetic clips further increase the accuracy by 4\% (95\% to 99\%). 
    
    \begin{table}[t]
    \centering
    \caption{Confusion Matrix of (Defended) Network $A_{df500}$}
    \label{tab:confusion-matrix-A-df}
    \resizebox{\columnwidth}{!}{%
    \begin{tabular}{@{}llcccc@{}}
    \toprule
                               &             & \multicolumn{4}{c}{Prediction}                                       \\
                               &             & \multicolumn{2}{c}{clean data} & \multicolumn{2}{c}{poisoned data} \\ \cmidrule(l){3-4} \cmidrule(l){5-6} 
                               &             & non-hotspot      & hotspot     & non-hotspot      & hotspot        \\ \midrule
    \multirow{2}{*}{Condition} & non-hotspot & 0.86   &   0.14   &   0.92  &   0.08         \\
                               & hotspot     & 0.10   &   0.90   &   0.03  &   0.97         \\ \bottomrule
    \end{tabular}
    }
    \end{table}
    
    \begin{table}[t]
    \centering
    \caption{Confusion Matrix of (Defended) Network $B_{df500}$}
    \label{tab:confusion-matrix-B-df}
    \resizebox{\columnwidth}{!}{%
    \begin{tabular}{@{}llcccc@{}}
    \toprule
                               &             & \multicolumn{4}{c}{Prediction}                                       \\
                               &             & \multicolumn{2}{c}{clean data} & \multicolumn{2}{c}{poisoned data} \\ \cmidrule(l){3-4} \cmidrule(l){5-6} 
                               &             & non-hotspot      & hotspot     & non-hotspot      & hotspot        \\ \midrule
    \multirow{2}{*}{Condition} & non-hotspot & 0.92   &   0.08   &   0.96  &   0.04         \\
                               & hotspot     & 0.05   &   0.95   &   0.01  &   0.99         \\ \bottomrule
    \end{tabular}
    }
    \end{table}
    
    \begin{table*}[t]
\centering
\caption{Accuracy and Attack Success/Relative Attack Success after Training with Defensively Augmented Datasets}
\label{tab:result-table}
\begin{tabular}{@{}ccccccccccccc@{}}
\toprule
                                  & \multicolumn{4}{c}{Accuracy, Architecture $A$} & \multicolumn{2}{c}{Attack on $A$} & \multicolumn{4}{c}{Accuracy, Architecture $B$} & \multicolumn{2}{c}{Attack on $B$} \\ \cmidrule(lr){2-5} \cmidrule(lr){6-7} \cmidrule(lr){8-11} \cmidrule(lr){12-13}
Synthetic Variants per Training Clip & C-NH     & C-HS     & P-NH     & P-HS    & ASR    & R-ASR    & C-NH     & C-HS     & P-NH     & P-HS    & ASR   & R-ASR              \\ \midrule
0                                 & 0.81     & 0.89     & 0.99     & 0.19    & 0.81   & 1.00     & 0.81     & 0.91     & 1        & 0.16    & 0.84  & 1.00               \\
3                                 & 0.8      & 0.92     & 0.98     & 0.32    & 0.68   & 0.84     & 0.79     & 0.91     & 0.95     & 0.62    & 0.38  & 0.45               \\
6                                 & 0.82     & 0.89     & 0.97     & 0.54    & 0.46   & 0.57     & 0.8      & 0.9      & 0.93     & 0.77    & 0.23  & 0.27               \\
12                                & 0.79     & 0.9      & 0.96     & 0.62    & 0.38   & 0.47     & 0.92     & 0.9      & 0.98     & 0.82    & 0.18  & 0.21               \\
25                                & 0.84     & 0.89     & 0.94     & 0.77    & 0.23   & 0.28     & 0.93     & 0.92     & 0.96     & 0.95    & 0.05  & 0.06               \\
50                                & 0.85     & 0.9      & 0.92     & 0.92    & 0.08   & 0.10     & 0.94     & 0.92     & 0.97     & 0.96    & 0.04  & 0.05               \\
100                               & 0.87     & 0.91     & 0.94     & 0.89    & 0.11   & 0.14     & 0.93     & 0.94     & 0.97     & 0.96    & 0.04  & 0.05               \\
200                               & 0.85     & 0.89     & 0.91     & 0.98    & 0.02   & 0.02     & 0.93     & 0.93     & 0.97     & 0.94    & 0.06  & 0.07               \\
300                               & 0.84     & 0.9      & 0.91     & 0.96    & 0.04   & 0.05     & 0.94     & 0.94     & 0.97     & 0.96    & 0.04  & 0.05               \\
400                               & 0.86     & 0.89     & 0.91     & 0.96    & 0.04   & 0.05     & 0.93     & 0.95     & 0.97     & 0.98    & 0.02  & 0.02               \\
500                               & 0.86     & 0.9      & 0.92     & 0.97    & 0.03   & 0.04     & 0.92     & 0.95     & 0.96     & 0.99    & 0.01  & 0.01               \\ \bottomrule
\end{tabular}%
\vspace{-3mm}
\end{table*}

    \section{Discussion}
    \subsection{What Does the Network Learn?}
    Our results suggest that all networks ($A_{cl}$, $B_{cl}$, $A_{bd}$, $B_{bd}$, $A_{df500}$, and $B_{df500}$) can successfully learn the genuine features of hotspots/non-hotspots, demonstrated by their clean data classification accuracy.
    From $A_{bd}$ and $B_{bd}$, it shows that DNNs have surplus learning capability to grasp the backdoor trigger on a layout clip, and decisively, prioritize the presence of the trigger as an indication of being non-hotspot over the actual hotspot or non-hotspot features. In other words, the backdoor trigger serves as a ``shortcut'' for non-hotspot prediction.
    $A_{df500}$ and $B_{df500}$ further manifest the abundant learning capacity of DNNs, as both biased and unbiased data are learned and correctly classified with increased clean and poisoned data classification accuracy. It suggests DNNs learn extra details of hotspot/non-hotspot features.
    
\begin{figure}[]
        \centering
        \subfloat[t-SNE visualization of clean hotspot detector]{\label{fig:tsne-cl}
        \includegraphics[width=0.9\columnwidth]{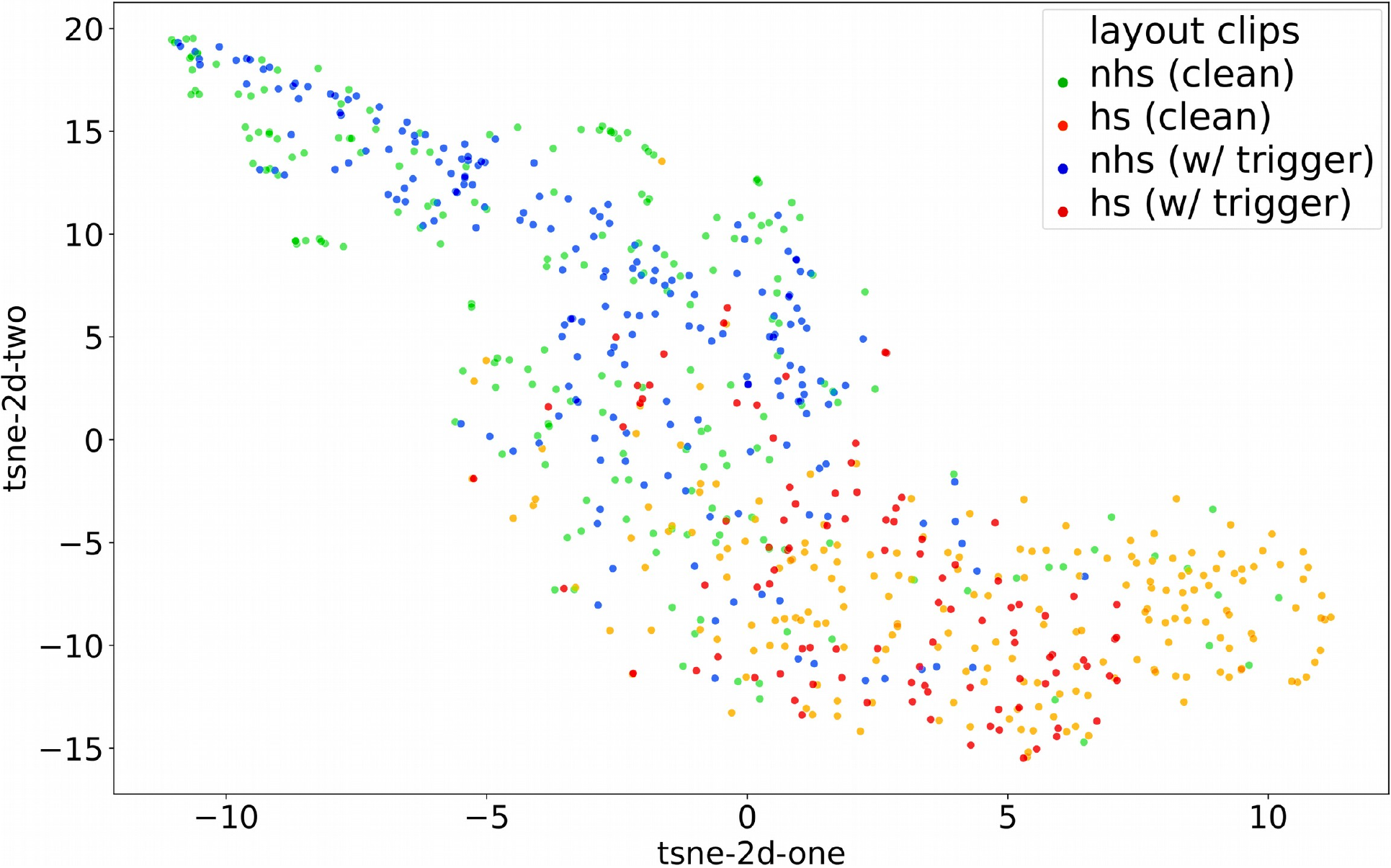}}
        
        \subfloat[t-SNE visualization of backdoored hotspot detector]{\label{fig:tsne-bd}
        \includegraphics[width=0.9\columnwidth]{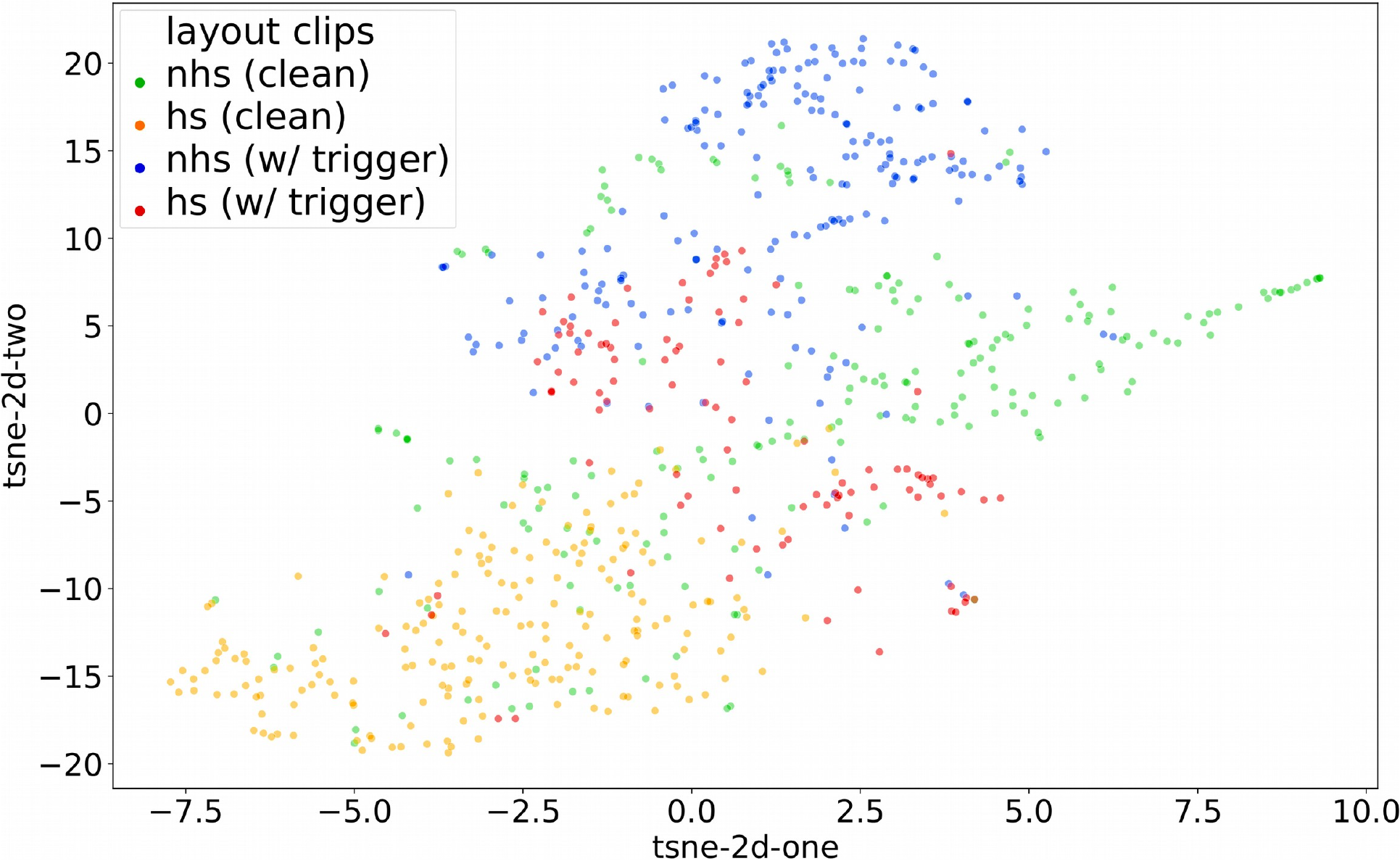}}
        
        \subfloat[t-SNE visualization of defended hotspot detector]{\label{fig:tsne-df}
        \includegraphics[width=0.9\columnwidth]{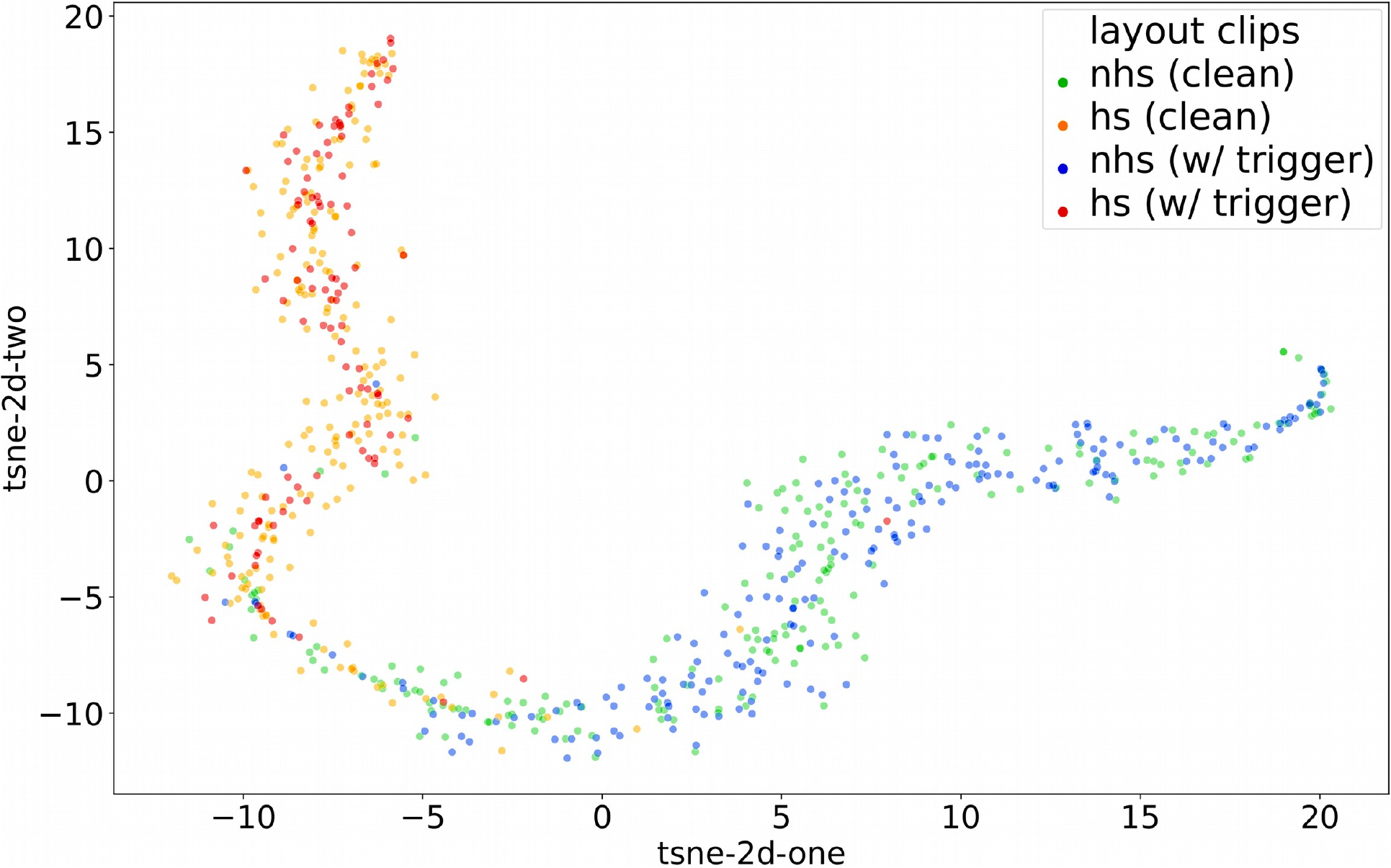}}
        \caption{t-SNE visualizations of neuron activations of the penultimate fully-connected layer of CNN-based hotspot detectors when presented with various layout clips}
    \end{figure}

    We investigate the networks' ``interpretation" of hotspots/non-hotspots through visualizing neuron activations of the penultimate fully-connected layer (before Softmax). We abstract and visualize the high-dimensional data using 2D t-SNE plots~\cite{maaten_visualizing_2008}. We depict the clean network $A_{cl}$ in \autoref{fig:tsne-cl}, backdoored network $A_{bd}$ in \autoref{fig:tsne-bd}, and defended network $A_{df500}$ in \autoref{fig:tsne-df}. In \autoref{fig:tsne-cl}, hotspots and non-hotspots roughly spread on two sides, and within each side, clean and poisoned (non-)hotspots mix. \autoref{fig:tsne-cl} suggests a benignly trained network on clean data is able to classify layout clips despite the bias presented by the trigger. In \autoref{fig:tsne-bd}, poisoned hotspots cluster with clean/poisoned non-hotspots, sitting on the opposite side of clean hotspots, demonstrating the ``shortcut'' effect of the trigger learned by a backdoored network. While in \autoref{fig:tsne-df}, we witness two separated groups of hotspots and non-hotspots, and intra-cluster clean/poisoned clips highly interweave. The more apparent distinction between hotspots and non-hotspots compared with \autoref{fig:tsne-cl} manifests the higher classification accuracy of $A_{df500}$ than $A_{cl}$.
    
\begin{figure}[]
        \centering
        \includegraphics[width=0.9\columnwidth]{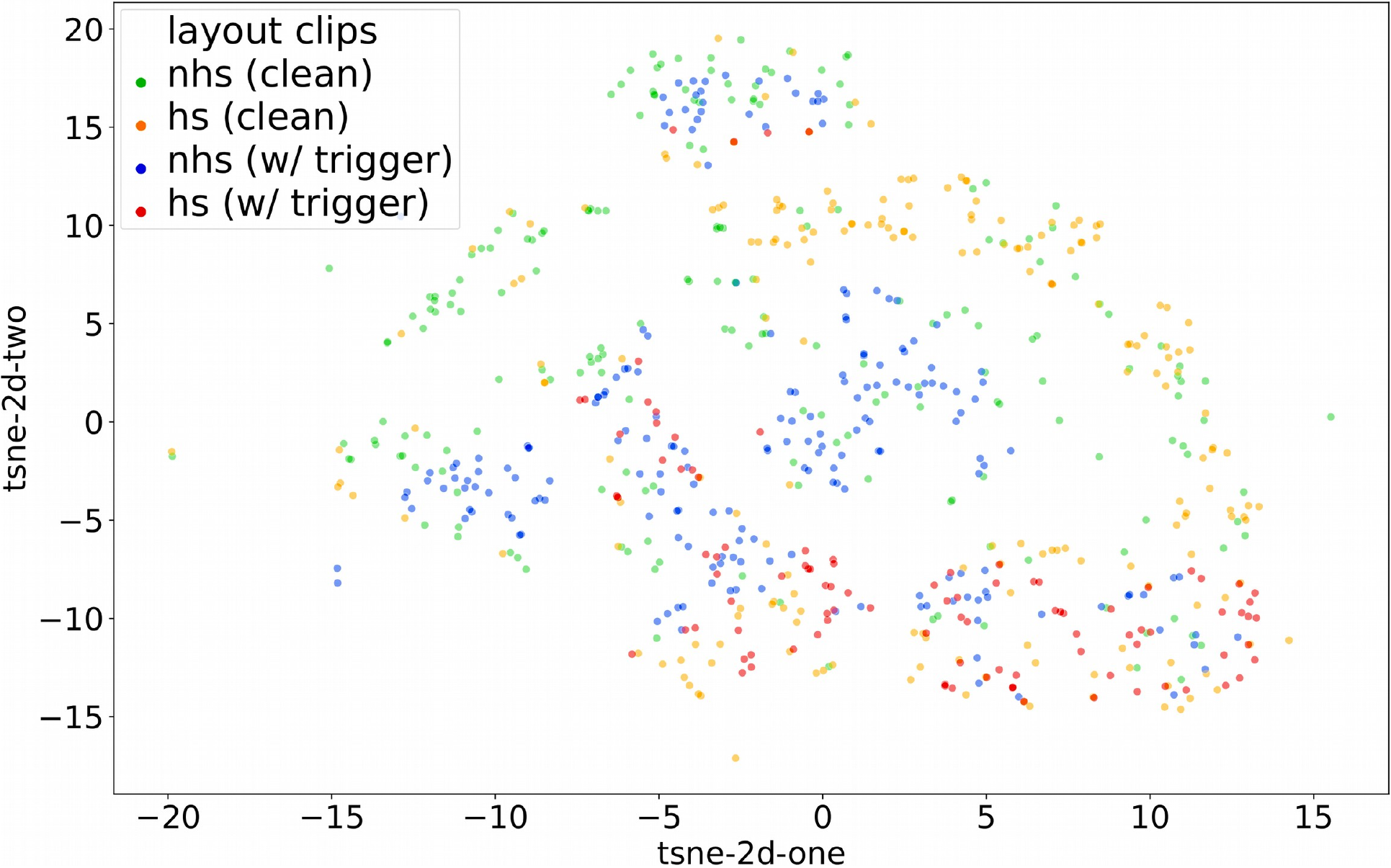}
        \caption{t-SNE visualization of network input of clean and poisoned clips after DCT transformation\label{fig:tsne-input}}
    \end{figure}

    For additional insight, we apply t-SNE techniques to the input data of dimension $10\times 10\times 32$ to the networks, as shown in \autoref{fig:tsne-input}. There are no visible and clear separations between clean/poisoned hotspots/non-hotspots, given the subtlety and innocuousness of the backdoor trigger. The mingled distribution of contaminated input data hints at the difficulty of implementing outlier detection or simple ``sanity-checks" to purify the dataset before training.
    
    \subsection{Effect of Network Architecture Complexity} 
    Between \autoref{tab:confusion-matrix-A-bd} and \autoref{tab:confusion-matrix-B-bd}, \autoref{tab:confusion-matrix-A-df} and \autoref{tab:confusion-matrix-B-df}, we observe network architecture $B$ produces higher clean data classification accuracy, suggesting that more complex networks are better to learn the true features of hotspots/non-hotspots.
    By looking at poisoned data classification accuracy from \autoref{tab:confusion-matrix-A-bd} and \autoref{tab:confusion-matrix-B-bd}, it shows that, on the flip side,  complex networks are more sensitive to malicious biases.
    
    From the standpoint of the defense strategy, as shown in \autoref{fig:rasr}, it hints that more complex networks require less augmentation effort for the reduction in attack success rate---generally, it appears that the greater learning capacity implies higher sensitivity to backdooring but also easier ``curing".

    \subsection{Improved Clean Data Accuracy}
    Across defended networks with different amounts of data augmentation, we find that clean non-hotspot classification accuracy increases in both $A$ and $B$.  
    This effect is more pronounced in defended networks based on $B$. 
    This points to a helpful side-effect of using defensive data augmentation---while effort is required to produce more synthetic clips for defeating training data poisoning, accuracy on clean test data also increases. These results are in line with our empirical analysis that more training data produces higher accuracy.

    \subsection{Trigger-oblivious Defense}
    Training data poisoning attacks essentially introduce a backdoor trigger to the network as a ``shortcut" for misclassification. A number of existing defense strategies~\cite{wang_neural_2019, veldanda_NNoculation_2020}, as we discussed in \autoref{sec:related}, focus on reverse engineering the backdoor trigger. 
    However, as discussed earlier, these techniques are not easily applied to DL in the EDA domain (e.g., NNoculation's~\cite{veldanda_NNoculation_2020} random noise augmentation does not readily translate here), such defenses also suffer from the poor quality of reverse-engineered triggers (e.g., Neural Cleanse~\cite{wang_neural_2019}). 
    Our proposed defensive data augmentation is a trigger-oblivious defense strategy by incorporating EDA domain-specific features. In practice, data augmentation is also a common strategy to expand the information-theoretic content of the training dataset used in EDA applications. Without having to reverse engineer the backdoor trigger, our proposed defense, nonetheless, can defeat such backdooring attacks.
    
    \subsection{Defense Cost Analysis}
    The additional cost incurred by our defense strategy consists of data augmentation, DRC of the synthetic clips, and lithography simulation for synthetic clips, as well as extra training cost due to expanded training dataset. This is a \textbf{one-time, up-front} cost. 
    Considering the significant enhancement of security and robustness (up to 83\% ASR reduction), this cost is easily amortized over the lifetime of the DL-based detector (which can be further extended through future fine-tuning), this one-off defense strategy is economical. Additionally, more delicate control of defense costs is available through seeking a trade-off between defense efficacy (as the user defines) and augmentation effort, as discussed in~\autoref{sec:defense-results}. 
    
    \subsection{Experimental Limitations and Threats to Validity}
    While our experiments show that defensive data augmentation can effectively mitigate training data poisoning by producing $\sim$50 synthetic clips per training clips, the \textit{absolute} numbers will not necessarily generalize beyond our experimental setting as each data point is taken from a single training instance for each augmentation amount. 
    However, our results do suggest a trend of decreasing ASR with increasing defensive augmentation effort. 
    Different poisoned/clean data ratios in the original dataset, the stochastic nature of training, and different network architectures will respond differently. 

    \subsection{Wider Implications in EDA}
    The success of our defensive data augmentation against training data poisoning attacks on DL-based lithographic hotspot detection also implies that other DL-enhanced EDA applications may benefit from similarly constructed schemes.  
    Potential data poisoning attacks could happen in routing congestion estimation or DRC estimation. Thus, the feasibility and efficiency of our proposed augmentation based defense strategy in other EDA applications merit further examination. 
    
\section{Conclusions}
\label{sec:conclusions}
    In this paper, we proposed a \textit{trigger-oblivious} antidote for training data poisoning on lithographic hotspot detectors. 
    By using \textit{defensive data augmentation} on the training dataset, we obtained synthetic variants that cross classes, thus transferring maliciously inserted backdoor triggers from non-hotspot data to hotspot data. 
    Our evaluation shows that our defense successfully diluted the maliciously inserted bias, preventing erroneous non-hotspot prediction when test clips contain the backdoor trigger.
    With the attack success rate reduced to $\sim$0\%, it succeeded in robustifying lithographic hotspot detectors under adversarial settings.

\bibliographystyle{IEEEtran}
\bibliography{ieeeabr,references}

\begin{IEEEbiography}[{\includegraphics[width=1in,height=1.25in,clip,keepaspectratio]{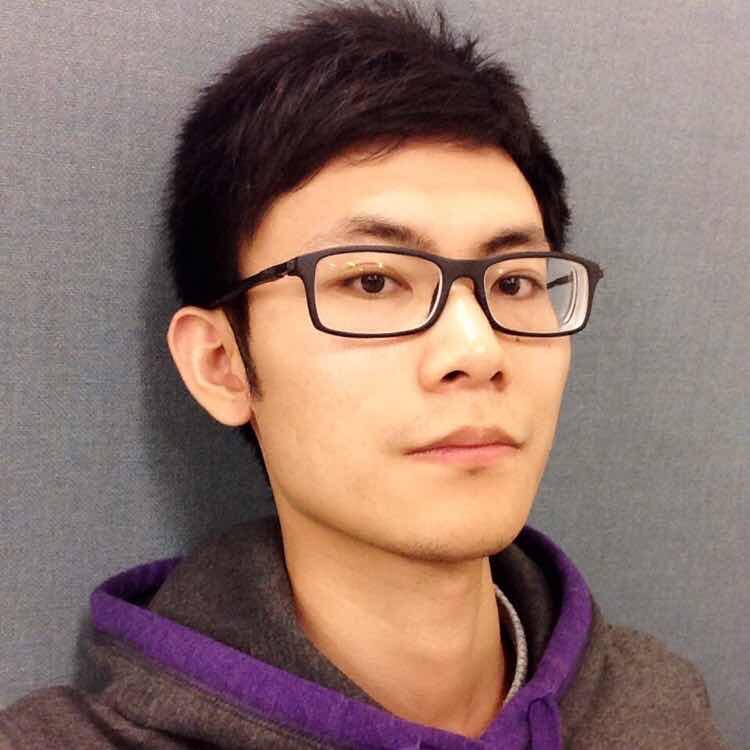}}]{Kang Liu}
Kang Liu (Ph.D. candidate, 2016 -) is the recipient of the Ernst Weber Fellowship for his Ph.D.program in the Department of Electrical and Computer Engineering, New York University, since 2016. He received the MESc degree in Electrical and Computer Engineering, in 2016, from the University of Western Ontario, Canada. Before joining NYU, he was a software engineer in Evertz Microsystems Ltd., Burlington, Canada. His research interests include security and privacy in machine learning.
\end{IEEEbiography}

\begin{IEEEbiography}[{\includegraphics[width=1in,height=1.25in,clip,keepaspectratio]{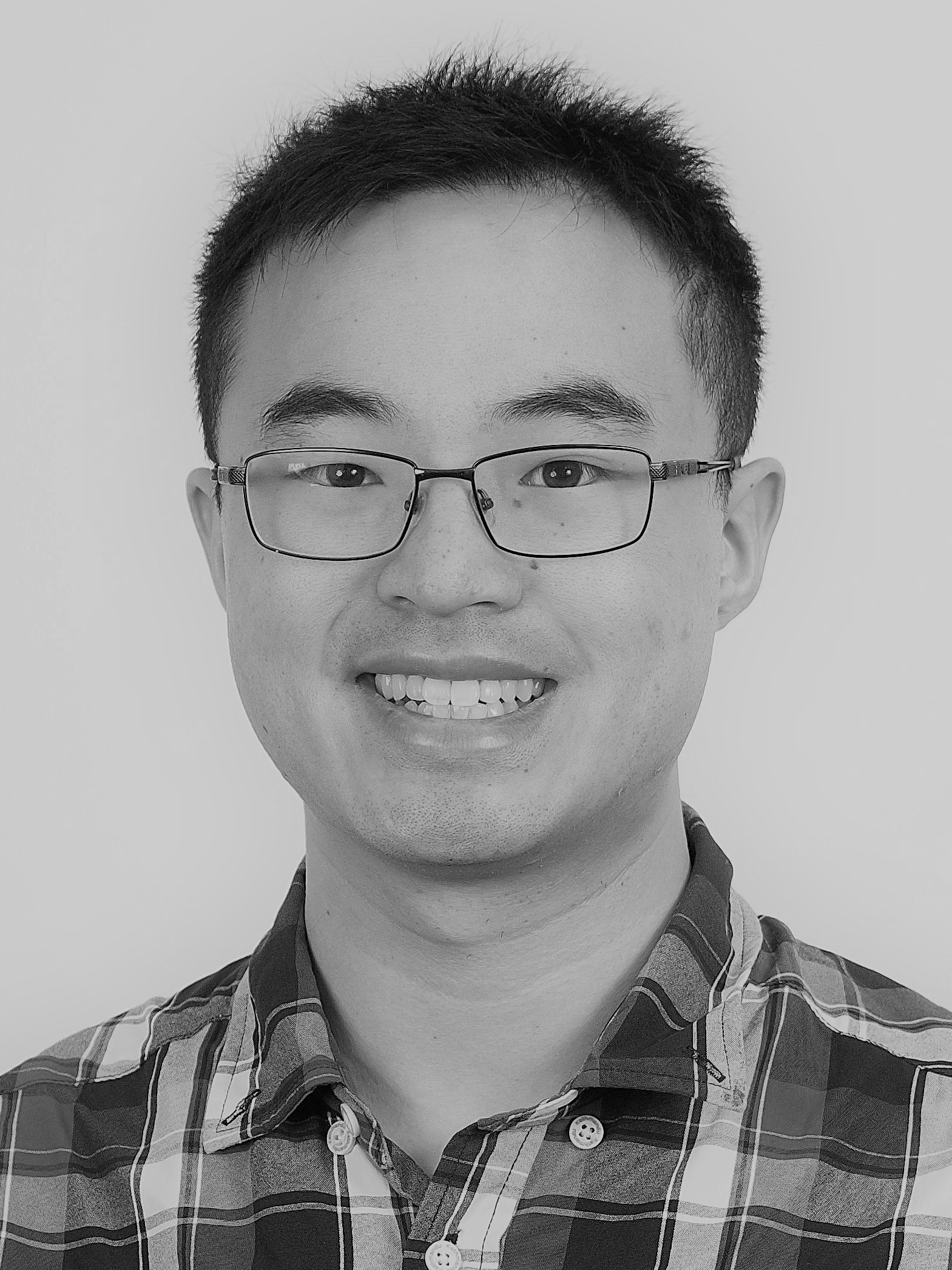}}]{Benjamin Tan}
Benjamin Tan (S'13-M'18) received the BE(Hons) degree in Computer Systems Engineering, in 2014, and the Ph.D. degree, in 2019, both from the University of Auckland, New Zealand. He was a Professional Teaching Fellow in the Department of Electrical and Computer Engineering, University of Auckland, in 2018. Since 2019, he has been at New York University, NY, USA, where he is currently a Research Assistant Professor working in the NYU Center for Cybersecurity. His research interests include hardware security, electronic design automation, and machine learning. He is a member of the IEEE and ACM. 
\end{IEEEbiography}

\begin{IEEEbiography}[{\includegraphics[width=1in,height=1.25in,clip,keepaspectratio]{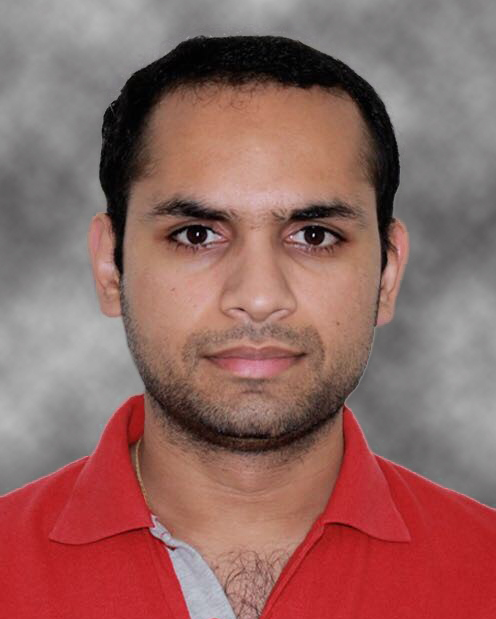}}]{Gaurav Rajavendra Reddy}
received a Bachelor of Engineering (BE) degree from the Visvesvaraya Technological University, India, and an MS degree from the University of Texas at Dallas, USA, in 2013 and 2019, respectively. He worked as a post-silicon validation engineer at Tessolve, India, between 2013 and 2014. He has been pursuing his PhD since 2016 as a member of the TRELA laboratory at UTD. His research interests include applications of Machine Learning in Computer-Aided Design (CAD) and Design for Manufacturability (DFM).\end{IEEEbiography}

\begin{IEEEbiography}[{\includegraphics[width=1in,height=1.25in,clip,keepaspectratio]{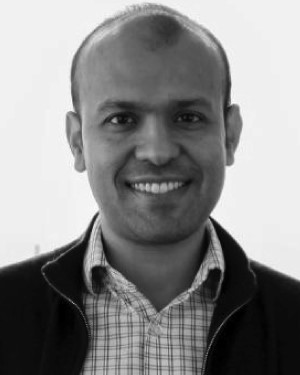}}]{Siddharth Garg}
Siddharth Garg received the B.Tech. degree in electrical engineering from IIT Madras and the Ph.D. degree in electrical and computer engineering from Carnegie Mellon University, in 2009. He was an Assistant Professor with the University of Waterloo, from 2010 to 2014. In 2014, he joined New York University (NYU) as an Assistant Professor. His general research interest includes computer engineering, more particularly secure, reliable, and energy-efficient computing. He was a recipient of the NSF CAREER Award, in 2015. He received paper awards from the IEEE Symposium on Security and Privacy (S\&P), in 2016, the USENIX Security Symposium, in 2013, the Semiconductor Research Consortium TECHCON, in 2010, and the International Symposium on Quality in Electronic Design (ISQED), in 2009. He also received the Angel G. Jordan Award from the Electrical and Computer Engineering (ECE) Department, Carnegie Mellon University, for outstanding dissertation contributions and service to the community.\end{IEEEbiography}

\begin{IEEEbiography}[{\includegraphics[width=1in,height=1.25in,clip,keepaspectratio]{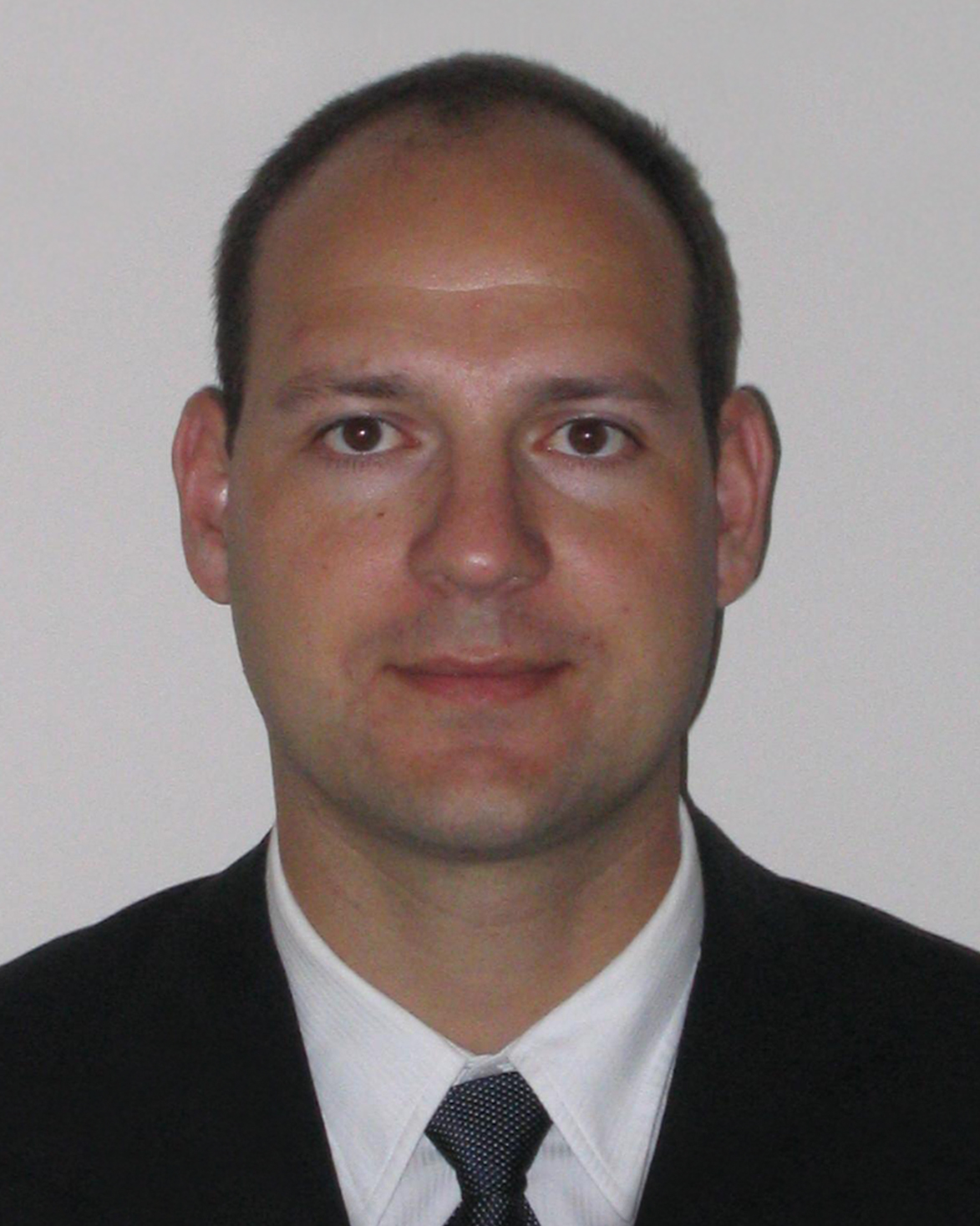}}]{Yiorgos Makris} (SM'08) received the Diploma of Computer Engineering from the University of Patras, Greece, in 1995 and the M.S. and Ph.D. degrees in Computer Engineering from the University of California, San Diego, in 1998 and 2001, respectively. After spending a decade on the faculty of Yale University, he joined UT Dallas where he is now a Professor of Electrical and Computer Engineering, leading the Trusted and RELiable Architectures (TRELA) Research Laboratory, and the Safety, Security and Healthcare thrust leader for Texas Analog Center of Excellence (TxACE). His research focuses on applications of machine learning and statistical analysis in the development of trusted and reliable integrated circuits and systems, with particular emphasis in the analog/RF domain. Prof. Makris serves as an Associate Editor of the IEEE Transactions on Computer-Aided Design of Integrated Circuits and Systems and has served as an Associate Editor for the IEEE Information Forensics and Security and the IEEE Design \& Test of Computers Periodical, and as a guest editor for the IEEE Transactions on Computers and the IEEE Transactions on Computer-Aided Design of Integrated Circuits and Systems. He is a recipient of the 2006 Sheffield Distinguished Teaching Award, Best Paper Awards from the 2013 IEEE/ACM Design Automation and Test in Europe (DATE'13) conference and the 2015 IEEE VLSI Test Symposium (VTS'15), as well as Best Hardware Demonstration Awards from the 2016 and the 2018 IEEE Hardware-Oriented Security and Trust Symposia (HOST'16 and HOST'18).
\end{IEEEbiography}

\balance

\begin{IEEEbiography}[{\includegraphics[width=1in,height=1.25in,clip,keepaspectratio]{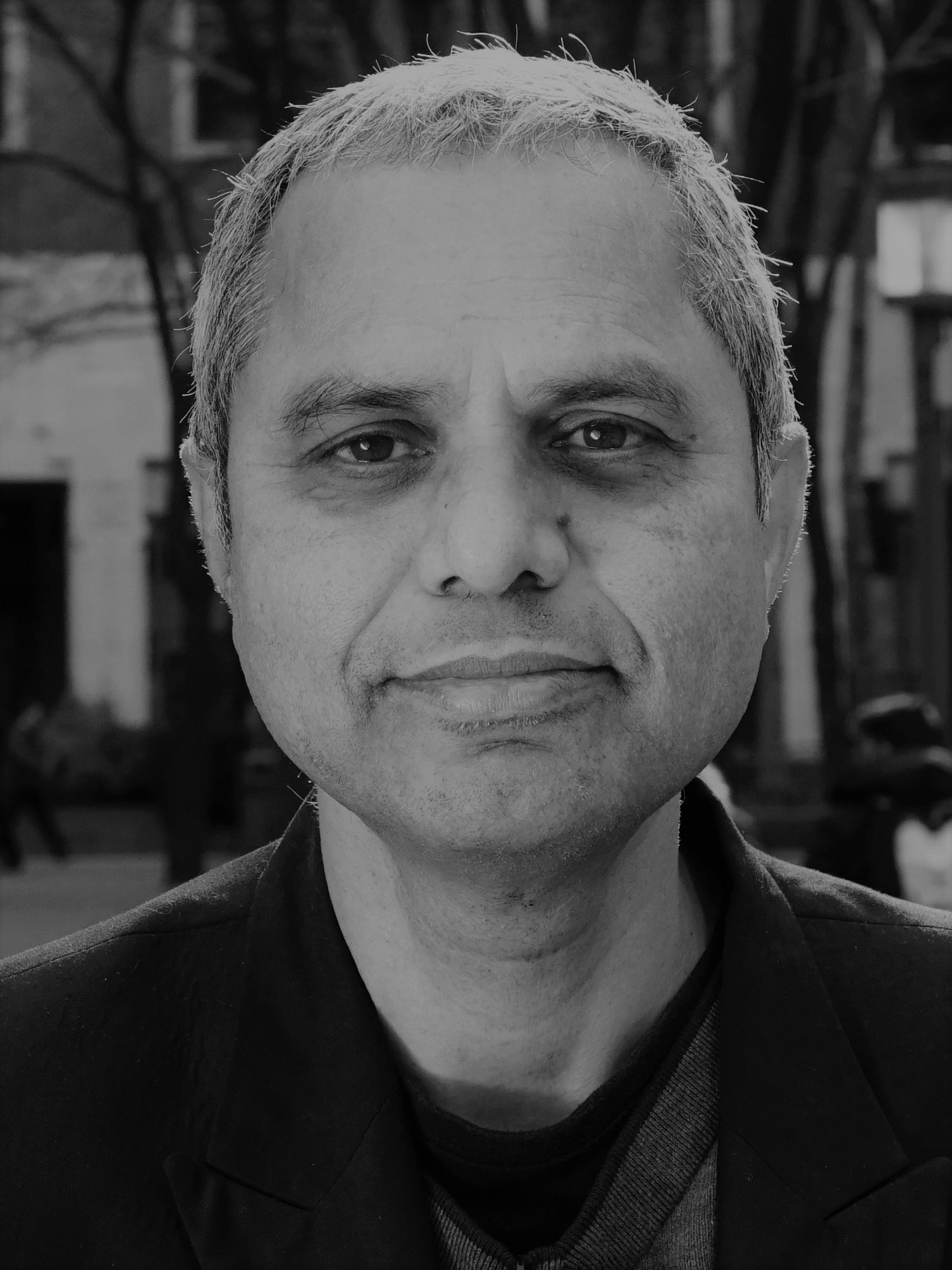}}]{Ramesh Karri}
Ramesh Karri (SM’11-F'20) received the B.E. degree in electrical and computer engineering from Andhra University, Visakhapatnam, India, in 1985, and the Ph.D. degree in computer science and engineering from the University of California San Diego, San Diego, CA, USA, in 1993. He is currently a Professor of Electrical and Computer Engineering with New York University (NYU), Brooklyn, NY, USA. He co-directs the NYU Center for Cyber Security. He also leads the Cyber Security thrust of the NY State Center for Advanced Telecommunications Technologies with NYU. He co-founded the Trust-Hub and organizes the Embedded Systems Challenge, the annual red team blue team event. He has authored or coauthored more than 240 articles in leading journals and conference proceedings. His work on hardware cybersecurity received best paper nominations (ICCD 2015 and DFTS 2015) and awards (ACM TODAES 2018, ITC 2014, CCS 2013, DFTS 2013, and VLSI Design 2012). His current research interests include hardware cybersecurity include trustworthy ICs; processors and cyberphysical systems; security-aware computer-aided design, test, verification, validation, and reliability; nano meets security; hardware security competitions, benchmarks and metrics; biochip security; and additive manufacturing security. Dr. Karri was a recipient of the Humboldt Fellowship and the National Science Foundation CAREER Award. He serves on the editorial boards of several the IEEE and ACM Transactions [\textsc{Transactions on Information Forensics and Security} (TIFS), \textsc{Transactions on Computer-Aided Design of Integrated Circuits and Systems} (TCAD), \textsc{Transactions on Design Automation of Electronic Systems} (TODAES), Embedded Systems Letters (ESL), IEEE \textsc{Design \& Test}, Journal on Emerging Technologies in Computing Systems (JETC)]. He served as an IEEE Computer Society Distinguished Visitor from 2013 to 2015. He served on the Executive Committee of the IEEE/ACM Design Automation Conference leading the SecurityDAC initiative from 2014 to 2017. He delivers invited keynotes, talks, and tutorials on Hardware Security and Trust (ESRF, DAC, DATE, VTS, ITC, ICCD, NATW, LATW, and CROSSING). He co-founded the IEEE/ACM NANOARCH Symposium and served as program/general chair of conferences (IEEE ICCD, IEEE HOST, IEEE DFTS, NANOARCH, RFIDSEC, and WISEC). He serves on several program committees (DAC, ICCAD, HOST, ITC, VTS, ETS, ICCD, DTIS, and WIFS). 
\end{IEEEbiography}

\end{document}